%% file: gnn.tex
\documentclass[sigconf]{acmart}

\usepackage{algorithmic}
\usepackage[utf8]{inputenc} 
\usepackage[T1]{fontenc}    
\usepackage{hyperref}       
\usepackage{url}            
\usepackage{booktabs}       
\usepackage{amsfonts}       
\usepackage{amsmath}
\usepackage{nicefrac}       
\usepackage{microtype}      
\usepackage{algorithm}
\usepackage{textcomp}
\usepackage{multirow}
\usepackage{numprint}
\usepackage[group-separator={,}]{siunitx}
\npthousandsep{,}\npthousandthpartsep{}\npdecimalsign{.}
\usepackage{setspace}

\usepackage{graphicx}
\usepackage{subcaption}

\newcommand{\CPU}{{\tt CPU}}
\newcommand{\GPU}{{\tt GPU}}

\newcommand{\zeroc}{{\tt 0c}}
\newcommand{\cdzero}{{\tt cd-0}}
\newcommand{\cdr}{{\tt cd-r}}
\newcommand{\cdf}{{\tt cd-5}}
\newcommand{\GNN}{{\tt GNN}}

\newcommand{\DL}{{\tt DL}}
\newcommand{\GAS}{{\tt GAS}}

\newcommand{\DGL}{{\tt DGL}}
\newcommand{\pyg}{{\tt PyG}}
\newcommand{\AP}{{\tt AP}}

\newcommand{\spmm}{{\tt SpMM}}
\newcommand{\GCN}{{\tt GCN}}
\newcommand{\MLP}{{\tt MLP}}
\newcommand{\DRPA}{{\tt DRPA}}
\newcommand{\DistGNN}{{\tt DistGNN}}
\newcommand{\DistDGL}{{\tt Dist-DGL}}
\newcommand{\RAT}{{\tt RAT}}
\newcommand{\LAT}{{\tt LAT}}
\newcommand{\NUMA}{{\tt NUMA}}

\AtBeginDocument{%
  \providecommand\BibTeX{{%
    \normalfont B\kern-0.5em{\scshape i\kern-0.25em b}\kern-0.8em\TeX}}}

\begin{document}

\title{DistGNN: Scalable Distributed Training for Large-Scale Graph Neural Networks}

\author{Vasimuddin Md, Sanchit Misra, Guixiang Ma, Ramanarayan Mohanty, Evangelos Georganas, Alexander Heinecke, Dhiraj Kalamkar, Nesreen K. Ahmed, Sasikanth Avancha}
\email{[vasimuddin.md, sanchit.misra, guixiang.ma, Ramanarayan.Mohanty, evangelos.georganas,}
\email{alexander.heinecke, dhiraj.d.kalamkar, nesreen.k.ahmed, sasikanth.avancha]@intel.com}
\affiliation{%
  \institution{Intel Corporation}
  \streetaddress{}
  \city{}
  \state{}
  \country{}
  \postcode{}
}


\renewcommand{\shortauthors}{}


\begin{abstract}

Full-batch training on Graph Neural Networks (\GNN{}) to learn the structure of large graphs is a critical problem that needs to scale to hundreds of compute nodes to be feasible. It is challenging due to large memory capacity and bandwidth requirements on a single compute node and high communication volumes across multiple nodes. 
In this paper, we present \DistGNN{} that optimizes the well-known Deep Graph Library (\DGL{}) for full-batch training on \CPU{} clusters via an efficient shared memory implementation, communication reduction using a minimum vertex-cut graph partitioning algorithm and communication avoidance using a family of delayed-update algorithms. Our results on four common \GNN{} benchmark datasets: Reddit, OGB-Products, OGB-Papers and Proteins, show up to $3.7\times$ speed-up using a single \CPU{} socket and up to $97\times$ speed-up using $128$ \CPU{} sockets, respectively, over baseline \DGL{} implementations running on a single \CPU{} socket.

\end{abstract}




\maketitle

\input{introduction}
\input{background}
\input{related_work}
\input{methods}
\input{results}
\input{conclusion}

\bibliographystyle{ACM-Reference-Format}
\bibliography{gnn}

\noindent{\small Optimization Notice: Software and workloads used in performance tests may have been optimized for performance only on Intel microprocessors. Performance tests, such as SYSmark and MobileMark, are measured using specific computer systems, components, software, operations and functions. Any change to any of those factors may cause the results to vary. You should consult other information and performance tests to assist you in fully evaluating your contemplated purchases, including the performance of that product when combined with other products. For more information go to
http://www.intel.com/performance. Intel, Xeon, and Intel Xeon Phi are trademarks of Intel Corporation in the U.S. and/or other countries.}

\end{document}

%% file: introduction.tex
\section{Introduction}
\label{sec-introduction}

Graphs are ubiquitous across multiple domains: social networks, power grids,  biological interactomes,  molecules etc.
In the fast-emerging domain of geometric deep learning~\cite{bronstein2017geometric}, a specific field called Graph Neural Networks (\GNN{}) has recently shown impressive results across a spectrum of graph and network representation learning problems~\cite{recommender, knowledgegraphs, drug-target, zitnik2018modeling}. \cite{powerfulgnn} shows that \GNN{}s are a very powerful mechanism to learn the structure of non-Euclidean data such as graphs. 
\GNN{}s combine low-dimensional embeddings associated with each vertex in a graph with local neighborhood connectivity for downstream machine learning analysis, e.g., vertex property prediction, link property prediction and graph property prediction. 

Scaling \GNN{} training for large graphs consisting of hundreds of millions of vertices and edges is a huge challenge, due to high memory capacity and bandwidth requirements as well as high communication volume to ensure convergence.
For large graphs, memory capacity requirements during training restrict the size of the problem that can be solved on a single socket. Mini-batch training works around these restrictions via neighborhood sampling~\cite{recommender, hamilton2017inductive, clustergcn, fastgcn}, to create a mini-batch of graph samples, which reduces the working set size. Another technique to mitigate memory capacity problems is to use the aggregate memory capacity of a distributed system~\cite{ucb, roc, neugraph}.
It has been shown that, in some cases, neighborhood sampling achieves lower accuracy compared to full-batch training~\cite{roc}. In this work, we focus on full-batch training and discuss distributed memory solutions that scale across mutliple Intel\textregistered~ Xeon\textregistered~ \CPU{} sockets; we plan to address scaling mini-batch training in future work.

\subsubsection*{Challenges of Full-batch Training:}
\GNN{} training poses the following specific challenges compared to traditional \DL{} workload training: (a)
Communication volume increases because vertex feature vectors and parameter gradients must be communicated, increasing pressure on the communications network, which could become a bottleneck;
(b) lower flops density and sequential nature of the training operations make it challenging to overlap computation with communication, exposing communication time;
(c) due to high byte-to-flop ratio, the aggregation operation tends to be memory and communication bandwidth bound.
On a single \CPU{} socket, optimal cache and memory bandwidth utilization are important factors in achieving good performance, while in a distributed setting, graph partitioning~\cite{libra, wb-libra, karypis1997metis} plays a crucial role in managing the communication bottleneck among compute sockets.

Current graph processing frameworks~\cite{powergraph, pregel, graphlab, graphx} are capable of processing large graphs, but they do not support \DL{} primitives. 
Consequently, the research community has developed libraries such as \DGL{}~\cite{dgl} and PyTorch Geometric (\pyg{})~\cite{pyg} with capability of message passing in the graphs.
These libraries employ current \DL{}-based frameworks such as TensorFlow~\cite{tensorflow}, PyTorch~\cite{pytorch}, and MXNet~\cite{mxnet}, coupling \DL{}  primitive operations with message passing.
However, a huge challenge is that key compute primitives within with these libraries are typically inefficient for single-socket shared-memory and multi-socket distributed-memory \CPU{} settings. We chose \DGL{} due to its rich functionality and flexibility. However, \DGL{}'s shared memory performance is poor on \CPU{}s and it does not support full batch training on distributed systems. In this work, we enhance \DGL{} by developing highly optimized single-socket and distributed solutions.


\subsubsection*{Present Work}
In this work, we discuss \DistGNN{}, which consists of a set of single-socket optimizations and a family of scalable, distributed solutions for large-scale \GNN{} training on clusters of Intel Xeon \CPU{}s. Our key contributions are as follows:

\begin{itemize}
\item A highly architecture-optimized implementation of multiple variants of the aggregation primitive implemented as a customized \spmm{} operation for single socket \CPU{}s. This is achieved by efficient cache blocking, dynamic thread scheduling and use of LIBXSMM~\cite{libxsmm} for loop reordering, vectorization and JITing to reduce instruction count. On a single \CPU{} socket (Intel Xeon 8280 with 28 cores), compared to the \DGL{} baseline, we achieve $3.66\times$ speedup for the GraphSAGE full-batch training on the Reddit dataset~\cite{hamilton2017inductive} (this is $3.1\times$ faster than what is reported in ~\cite{avancha2020deep}) and $1.95\times$ speedup on OGB-Products dataset~\cite{ogb}. 
\item Novel application of vertex-cut based graph partitioning to large-graphs to achieve optimal reduction of communication across \CPU{} sockets during full-batch \GNN{} training.
\item Novel application of delayed update algorithm to feature aggregation to achieve optimal communication avoidance during full-batch \GNN{} training.  To reduce the impact of communication, we overlap it with computation by spreading it across across epochs; overlap occurs at the expense of freshness -- aggregation uses stale, partially-aggregated remote vertex features. Our optimizations for single socket make it even more challenging and critical to overlap communication with computation.
\item First-ever demonstration of \textbf{full-batch} \GNN{} training on \CPU{}-based distributed memory systems. We showcase the performance of our distributed memory solution on the OGB-papers dataset~\cite{ogb}, which contains $111$ million vertices and $1.6$ billion edges. Our solution scales GraphSAGE full-batch training to $128$ Intel Xeon \CPU{} sockets, achieving $97\times$ speedup compared to the \DGL{} (un-optimized) baseline and $83\times$ compared to our optimized implementation, respectively, running on a single \CPU{} socket.
\end{itemize}

The paper is organized as follows. Section~\ref{sec-background} describes the core operation of \GNN{}, the aggregation function, and \DGL{} library.
Section~\ref{sec-related} positions the related literature. In Section~\ref{sec-distgnn-shared}, we discuss our shared memory solution. We analyze compute characteristics of \GNN{} operations and detail various architecture-aware optimizations. Section~\ref{sec-distgnn-distributed} describes graph partitioning and a set of distributed algorithms. Section~\ref{sec-results} demonstrates the performance of our solutions. Section ~\ref{sec-conclusion} summarizes our work and discusses future work.

%% file: background.tex
\section{Background}
\label{sec-background}

\GNN{}s learn graph structure through low-dimension embeddings of vertices or edges.
They compute these embeddings using an {\em aggregation function}, which recursively gathers multi-hop neighborhood features to encode vertex or edge features.
The aggregation function also learns the shared weights by training a shallow neural network on the gathered features.
Depending on the application, neighborhood aggregation precedes or succeeds the neural network layers.
The aggregation function operates via {\em message-passing} between vertices or/and edges using an {\em Aggregation Primitive} (\AP{}), which constitute a core part of the aggregation function.


\subsection{Aggregation Primitive}
\label{sec-primitives}

Let $\mathcal{G}(\mathcal{V}, \mathcal{E})$ be an input graph with vertices $\mathcal{V}$ and edges $\mathcal{E}$, and let $f_{\mathcal{V}}$ and $f_{\mathcal{E}}$ be the vertex and edge features, respectively. The sizes of $f_{\mathcal{V}}$ and $f_{\mathcal{E}}$ are $|\mathcal{V}| \times d$ and $|\mathcal{E}| \times d$, respectively, where $d$ is the feature vector size. 
In effect, the \AP{} is a tuple $(f_{\mathcal{V}}, f_{\mathcal{E}}, \otimes, \oplus, f_{\mathcal{O}})$, where $\otimes$ and $\oplus$ are element-wise operators on $f_{\mathcal{V}}$ or $f_{\mathcal{E}}$ (or a combination thereof) to produce output features $f_{\mathcal{O}}$. 

In \AP{}, operator $\otimes$ can be an element-wise binary or unary operator. In binary form, it operates on a pair of inputs; valid pairs are ($f_\mathcal{V}$, $f_\mathcal{V}$) and ($f_\mathcal{V}$, $f_\mathcal{E}$), in an appropriate order. Operator $\oplus$ acts as element-wise {\em reducer} that reduces the result of binary operation on to the final output. Mathematically, \AP{} can be expressed as,

\begin{gather}
    \AP{}(x, y, \otimes, \oplus, z) : \oplus(\otimes(x,y), z),
    \label{eq:br} \\
    \forall x, y, z \in f_\mathcal{V}, f_\mathcal{E} \nonumber
\end{gather}




If one of the inputs is {\tt NULL}, then operator $\otimes$ takes the unary form; it copies the input features and reduces them to the final output. Mathematically, assuming $y=\phi$, (where $\phi$ stands for NULL) \AP{} using the unary operator is,

\begin{gather}
    \AP{}(x, \phi, \otimes, \oplus, z) : \oplus(copy(x), z),
    \label{eq:cr} \\
    \forall x, y, z \in f_\mathcal{V}, f_\mathcal{E} \nonumber
\end{gather}




Given a graph, Equations~\ref{eq:br} and ~\ref{eq:cr} can be formulated as \spmm{}:  $f_O = A \times f_X$, where $A$ is the graph adjacency matrix and $f_X$ is the dense feature matrix. 
\DGL{} featgraph~\cite{featgraph} models the \AP{} primitive as \spmm{} and provides a single template API for \AP{} computations. 
In section~\ref{sec-distgnn-shared}, we describe variants of \spmm{} for different forms of \AP{} and rigorous architecture-aware optimizations to accelerate it.

\subsection{GNN Libraries}

\DGL{} and \pyg{} are the two prominent emerging libraries for performing \GNN{} operations. 
Due to its rich set of features, in this work, we use \DGL{} to implement our single-socket and distributed algorithms.

\DGL{} provides graph abstractions with rich set of functionality for manipulation and utilization of graph objects.
It filters out general computations on graphs as the message passing paradigm.
For computations on vertices, the message-passing functionality is formulated as \spmm{}.
For computations on edges, the message-passing functionality is formulated as sampled dense-dense matrix multiplication ({\tt SDDMM}).
It provides in-built support for various binary and reduction operators as well as support for user-defined functions.
The distillation of the core \GNN{} operations as a few matrix multiplication operations enables parallelization and various other optimizations, such as vectorization.
Additionally, \DGL{} relies on popular \DL{} frameworks for its neural network operations. 
It provides the flexibility of selecting from popular backend \DL{} frameworks such as TensorFlow, PyTorch, and MXNet.

%% file: related_work.tex
\section{Related Work}
\label{sec-related}

Prior art in this field encompasses both full-batch and mini-batch training approaches. Due to small model sizes relative to other \DL{} workloads, distributed \GNN{}s employ data parallelism, by partitioning the input graph across \CPU{} sockets or \GPU{} cards. 

A few approaches have been proposed for GPU-based distributed memory systems, which we describe briefly below.
NeuGraph~\cite{neugraph} describes a single node multi-GPU parallel system for training large-scale \GNN{}.
It introduces a new programming model for \GNN{}, leveraging a variant of vertex-centric parallel graph abstraction \GAS{}~\cite{powergraph}.
NeuGraph partitions the input graph using min-cut Metis partitioning. 
Roc~\cite{roc}, a distributed multi-GPU \GNN{} training system, applies an online regression model to get the graph partitions.
Coupled with sophisticated memory management between the host and the \GPU{} and a fast graph propagation optimized \GPU{} tool called Lux,
Roc showcases scalable performance for large graphs on distributed \GPU{} system. Leveraging their scalable solution, Roc trains more complex \GNN{} model architecture to achieve better model accuracy. Roc demonstrates the efficacy of their solution on small benchmark datasets of Reddit and Amazon.
CAGNET~\cite{ucb} implemented a suite of parallel algorithms: 1D, 1.5D, 2D, and 3D algorithms, for \GNN{} training using complete neighborhood aggregation.
Inspired by SUMMA algorithm~\cite{summa} for matrix multiplication, they apply different kinds of matrix blocking strategies for work division among the compute nodes.
Their solution uses optimized cuSPARSE for the  \spmm{} computations; however, it suffers from poor scaling due to communication bottlenecks.

For CPU-based distributed systems, all the approaches proposed so far have been for mini-batch training as described below.
AliGraph~\cite{aligraph} presents a comprehensive framework building various \GNN{} applications. It supports distributed storage, sampling, and aggregator operators,
along with a suite of graph partitioning techniques including vertex- and edge-cut based ones.
AliGraph implements and showcases the efficacy of the interesting concept of caching the neighbors of important vertices to reduce the communication load.
However, AliGraph does not report the scaling of their solution. 
\DistDGL{}~\cite{distdgl}, a distributed \GNN{} processing layer added to the \DGL{} framework.
It holds the vertex features in a distributed data server which can be queried for data access.
Due to the large execution time of their inefficient graph sampling operation, it manages to overlap communication with the sampling time. \DistDGL{} demonstrates linear scaling on the largest benchmark dataset, the OGBN-Papers.

Various graph processing frameworks for graph analytics and machine learning have been proposed in the literature~\cite{pregel, graphlab, powergraph, graphx}. 
The most notable property of these frameworks is the Gather-Apply-Scatter (\GAS{}) model.
The aggregate step in \GNN{} is a simple message passing step between the neighbors with synchronization at the end of the step.
These frameworks using synchronous \GAS{} can naturally implement the aggregate step of \GNN{}; however, they lack the support for implementing various graph neural network operations as well as the graph attention models.

%% file: methods.tex




\section{DistGNN: Shared-Memory Algorithm}
\label{sec-distgnn-shared}

\AP{} accounts for a majority of the run-time in \GNN{} applications. In this section, we describe techniques we have created to accelerate their implementations within \DGL{} for a shared memory system. 

\vspace{-10pt}
\subsection{Baseline Implementation of Aggregation Primitive in \DGL{}}
Alg. \ref{algo:ap-dgl} provides a pseudo code of the \AP{} in \DGL{}. It uses a customized SpMM-like formulation. Given an input graph $\mathcal{G}(\mathcal{V}, \mathcal{E})$, let $A$ be its adjacency matrix in CSR format. $A[v]$ gives the list of neighbors of $v$ from which there is an edge incident on $v$. For each edge $e_{uv} \in \mathcal{E}$, we call $u$ the source vertex and $v$ the destination vertex. \DGL{} pulls messages from vertex $u$ and edge $e_{uv}$ and reduces them into vertex $v$. More formally, for each edge $e_{uv}$ incident from vertex $u$ to vertex $v$, \DGL{} computes the binary operator $\otimes$ element-wise between corresponding vertex feature vector of $u$ ($f_\mathcal{V}[u]$) and edge feature vector of $e_{uv}$ ($f_\mathcal{E}[e_{uv}]$). Subsequently, it reduces the result with the vertex features of $v$ in the output feature matrix ($f_\mathcal{O}[v]$) by computing the reduction operator $\oplus$ element-wise and stores it into $f_\mathcal{O}[v]$. $\otimes$ can also be a unary operator. In that case, $\otimes$ is computed on either $f_\mathcal{V}[u]$ or $f_\mathcal{E}[e_{uv}]$ and reduced with $f_\mathcal{O}[v]$. Table ~\ref{table:dgl-op} details various binary/unary operators ($\otimes$) and reduction operators ($\oplus$) used in \DGL{}. 

The computation is parallelized by distributing destination vertices across OpenMP threads. This way, there are no collisions because only one thread \emph{owns} the feature vector $f_\mathcal{O}[v]$ at each destination vertex  ($v$) and reduces the \emph{pulled} source feature vectors into $f_\mathcal{O}[v]$. 

\begin{algorithm}[htbp]
\caption{Aggregation Primitive in \DGL{}}
\label{algo:ap-dgl}
\begin{algorithmic}[1]
\small
\REQUIRE Matrix $A$ of size $|\mathcal{V}| \times |\mathcal{V}|$ in CSR format
\REQUIRE Input feature matrix $f_{\mathcal{V}}$ of size $ |\mathcal{V}| \times d$ (Vertex feature set)
\REQUIRE Input feature matrix $f_{\mathcal{E}}$ of size $ |\mathcal{E}| \times d$ (Edge feature set)
\REQUIRE Output feature matrix $f_\mathcal{O}$ of size $ |\mathcal{V}| \times d$ (Initialized to zeros)
\REQUIRE Unary/Binary operator: $\otimes$, Reduction operator: $\oplus$
\FOR{ $v \in \mathcal{V}$ in parallel }
    \FOR{ $u$ in $A[v]$ }
        \STATE $f_{\mathcal{O}}[v] \leftarrow f_{\mathcal{O}}[v] \oplus (f_{\mathcal{V}}[u] \otimes f_{\mathcal{E}}[e_{uv}])$
    \ENDFOR
\ENDFOR
\end{algorithmic}                            
\end{algorithm}

\begin{table}[htbp]
\caption{Binary/Unary and Reduction Operators in \DGL{}}
\vspace{-10pt}
\label{table:dgl-op}
\begin{tabular}{ccccc}
\hline
Unary/Binary & operands & Reduction &output ($z$) \\
Operator ($\otimes$) &  & Operator ($\oplus$) & \\
\hline
add/sub/mul/div  & $x$,$y$ & sum/max/min & $z \oplus (x \otimes y)$ \\
copylhs  & $x$, $\phi$ & sum/max/min & $z \oplus ($copy$(x))$ \\
copyrhs  & $\phi$, $y$ & sum/max/min & $z \oplus ($copy$(y))$ \\
\hline
\end{tabular}
\end{table}

\subsection{Efficient Aggregation Primitive}

In case of large graphs, the feature matrices -- $f_{\mathcal{V}}$, $f_{\mathcal{E}}$ and $f_{\mathcal{O}}$ -- don't fit in cache. Moreover, real world graphs are typically highly sparse. To aggregate the feature vector of a vertex, the feature vectors of all its neighbors and the corresponding edges must be accessed. Feature vectors of edges incident on a vertex can be contiguously stored in $f_{\mathcal{E}}$ and are only accessed once in Alg. ~\ref{algo:ap-dgl}. This effectively makes it a contiguous access of large enough block of memory that is used once  - hence, a memory bandwidth (BW) bound streaming access. On the other hand, those of the neighbors in $f_{\mathcal{V}}$ can be non-contiguous, sparsely located (leading to random gathers) and will be used as many times as the number of their neighbors. Moreover, a feature vector, $f_\mathcal{V}[v]$, accessed once and brought into cache may get thrashed out before it is needed again. Hence, in many cases, $f_\mathcal{V}[v]$ needs to be fetched from memory despite having been accessed before. Each such access adds to memory BW requirement and also has to pay the memory latency cost for the first few cache lines before the hardware prefetcher kicks in. 

\subsubsection*{Cache blocking}
Therefore, we need to apply cache blocking to take advantage of the {\it cache reuse} present in the graph, and avoid random gathers. We can either block $f_\mathcal{V}$ and run through entire $f_\mathcal{O}$ for every block of $f_\mathcal{V}$ or block $f_\mathcal{O}$ and run through entire $f_\mathcal{V}$ for every block of $f_\mathcal{O}$. In the latter case, we need to parallelize across source vertices ($u$) leading to race conditions on destination vertices ($v$). Therefore, we block $f_\mathcal{V}$. Alg. ~\ref{algo:ap-blocked} illustrates how we apply cache blocking using blocks of size $B$. First, we create $n_B$ blocks by creating a CSR matrix for each block for easy access of neighbors. For each block, we go through all the destination vertices ($v$) in parallel ensuring that all threads work on one block of $B$ source vertices at a time. As a result, any feature vector in $f_\mathcal{V}$ read by some thread $t$ could be in the L2 cache of the \CPU{} if/when some other thread $t'$ reads the same feature vector. Therefore, $f_\mathcal{V}$ is read from memory only once but we make $n_B$ passes over $f_\mathcal{O}$. Each additional pass of $f_\mathcal{O}$ adds to BW requirement. Hence, the block size ($B$) should be as large as possible while ensuring that a block of $f_\mathcal{V}$ can fit into cache. Sparser graphs allow us to have larger $B$ as not all feature vectors of $f_\mathcal{V}$ are active in a block. However, finding the best block size is challenging since many graphs follow a power law and there could be vertices with extremely large neighborhoods resulting in more feature vectors being active in the corresponding block of $f_\mathcal{V}$. 


\begin{algorithm}[htbp]
\caption{Application of Blocking on Aggregation Primitive}
\label{algo:ap-blocked}
\begin{algorithmic}[1]
\small
\REQUIRE Matrix $A$ of size $|\mathcal{V}| \times |\mathcal{V}|$ in CSR format
\REQUIRE Input feature matrix $f_{\mathcal{V}}$ of size $ |\mathcal{V}| \times d$ (Vertex feature set)
\REQUIRE Input feature matrix $f_{\mathcal{E}}$ of size $ |\mathcal{E}| \times d$ (Edge feature set)
\REQUIRE Output feature matrix $f_{\mathcal{O}}$ of size $ |\mathcal{V}| \times d$ (Initialized to zeros)
\REQUIRE Unary/Binary operator: $\otimes$, Reduction operator: $\oplus$
\REQUIRE Block size, $B$
\STATE $n_B \leftarrow \left\lceil\frac{|V|}{B}\right\rceil$ \COMMENT{Number of blocks}
\STATE $\{A_0, A_1, \ldots, A_{n_B - 1}\} \leftarrow$ Create CSR matrices for all blocks
\FOR{ $i \in 0,\ldots,n_B - 1$ }
    \FOR{ $v \in \mathcal{V}$ in parallel }
        \FOR{ $u$ in $A_i[v]$ }
            \STATE $f_{\mathcal{O}}[v] \leftarrow f_{\mathcal{O}}[v] \oplus (f_{\mathcal{V}}[u] \otimes f_{\mathcal{E}}[e_{uv}])$
        \ENDFOR
    \ENDFOR
\ENDFOR
\end{algorithmic}                            
\end{algorithm}

\vspace{-10pt}
\subsubsection*{Multithreading}
Many graphs following the power law could result in a significant difference in number of neighbors across vertices. Therefore, we use dynamic thread scheduling with OpenMP, allocating a chunk of contiguous destination vertices at a time to ensure that a thread writes to contiguous feature vectors of $f_\mathcal{O}$.

\subsubsection*{Loop reordering and Vectorization with LIBXSMM}
Applying SIMD to the innermost loop of Alg. ~\ref{algo:ap-blocked} (lines \#5-7) for the large variety of binary/unary and reduction operators in \DGL{} using manually written intrinsics could be a time consuming task. Instead, we use the LIBXSMM library~\cite{libxsmm} that provides highly architecture optimized primitives for many matrix operations including our use-cases. LIBXSMM reorders the loop (Alg. ~\ref{algo:ap-lr}) to ensure each $f_{\mathcal{O}}[v]$ is written to only once per block. It generates optimal assembly code with SIMD intrinsics where applicable using JITing thus providing more instruction reduction than manually written intrinsics based code.

\begin{algorithm}[htbp]
\caption{Reordering of the loop at lines \#5-7 in Alg. ~\ref{algo:ap-blocked}}
\label{algo:ap-lr}
\begin{algorithmic}[1]
\small
\REQUIRE SIMD Width, $W$
        \FOR{ $j \in 0,\ldots,d-1$, step $W$ }
        \STATE $t \leftarrow f_{\mathcal{O}}[v][j:j+W-1]$ 
        \FOR{ $u$ in $A_i[v]$ }
            \STATE $t \leftarrow t \oplus (f_{\mathcal{V}}[u][j:j+W-1] \otimes f_{\mathcal{E}}[e_{uv}][j:j+W-1])$
        \ENDFOR
        \STATE $f_{\mathcal{O}}[v][j:j+W-1] \leftarrow t$
        \ENDFOR
\end{algorithmic}                            
\end{algorithm}

\section{DistGNN: Distributed-Memory Algorithm}
\label{sec-distgnn-distributed}

In this section, we describe our distributed algorithms using GraphSAGE \GNN{} model.
As seen in the previous sections, feature aggregation operation is the dominant runtime component. \GNN{} models consist of a relatively small enough number of parameters, which can be processed on a single socket. However, with the increase in graph size, the aggregation operation is limited by the available memory. Hence, our distributed parallel solutions use data-parallelism. The model, being smaller in size, is replicated on the sockets and the input graph is partitioned. 

It is desirable that partitions communicate less frequently during aggregation. Ideally, partitions would be fully self-contained and require no communication with each other; however, this is not practical and will result in lower training accuracy. In practice, we partition the graph to minimize communication.

\subsection{Graph Partitioning}

Real-world graphs follow power-law degree distribution. ~\cite{percolation} shows that vertex-cut produces minimal cuts for power-law graphs. 
Distributed graph processing frameworks have shown the efficacy of vertex-cut based partitioning~\cite{powergraph}. In this work, we use vertex-cut based graph partitions. 
These partitioning techniques distribute the edges among the partitions. Thus, each edge is present in only one partition, while a vertex can reside in multiple ones. Each vertex that splits due to the vertex-cut carries with it a partial neighborhood of the original vertex from the input graph. Thus, any update to such a vertex must be communicated to its clones in other partitions. The number of such clones (for each original vertex that is split) is called {\em replication factor}.
The goal of the partitioning is two-fold: (a) to reduce  communication across partitions and (b) generate balanced partitions. 
The first goal can be achieved by maintaining a lower replication factor. Towards the second goal, we use uniform edge distribution as the load balancing metric.
In this work, we use a state-of-the-art tool called {\em Libra}~\cite{libra} to partition our un-weighted input graph. 
Libra works on a simple principle for graph partitioning. It partitions the edges by assigning them to the least-loaded relevant (based on edge vertices) partition. For \GNN{} benchmark datasets, we observe that Libra produces balanced partitions in terms of edges.  

\begin{figure}[ht]
\begin{subfigure}{0.43\linewidth}
\includegraphics[width=0.9\linewidth]{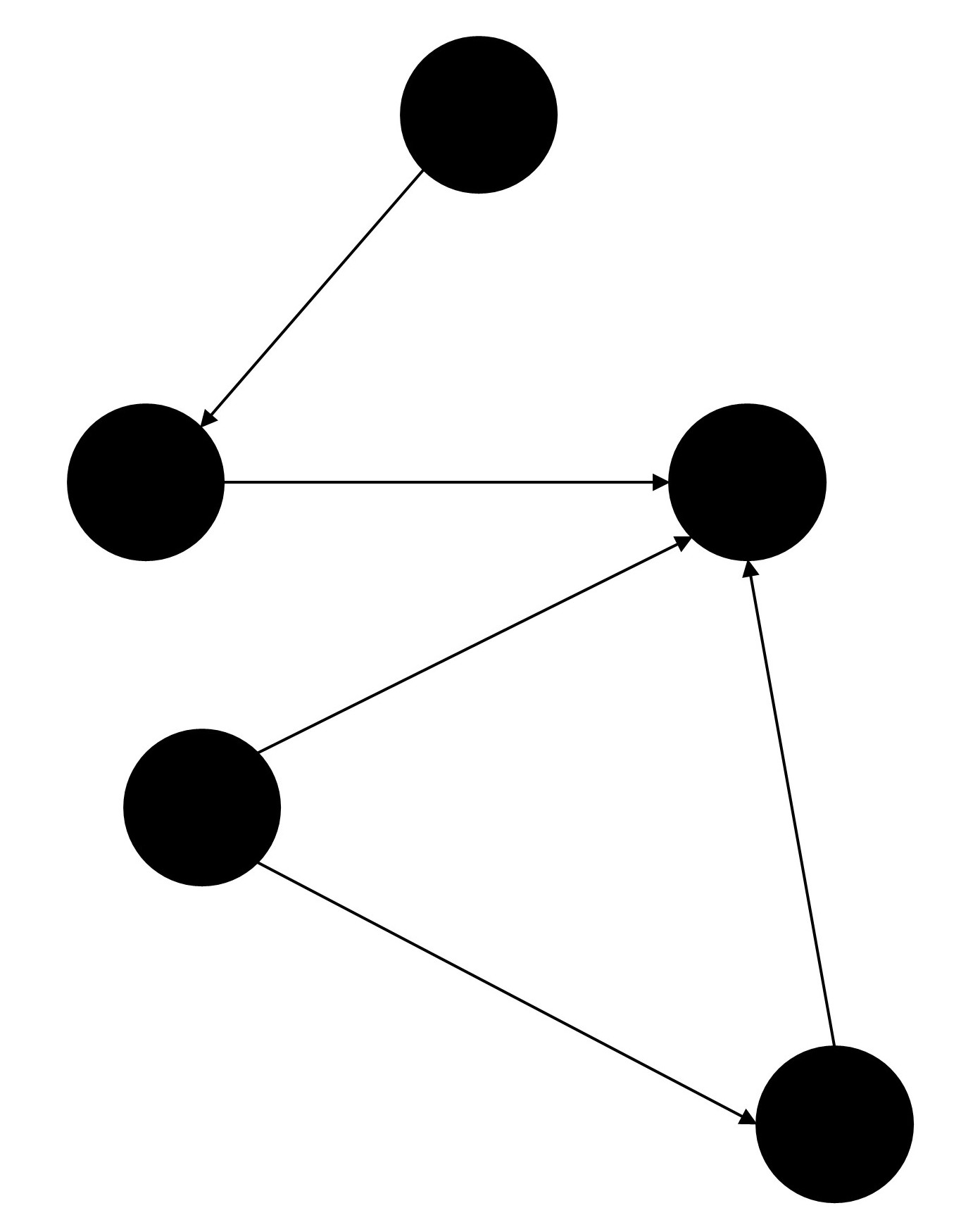}
\caption{Original graph}
\end{subfigure}
\begin{subfigure}{0.56\linewidth}
\includegraphics[width=0.9\linewidth]{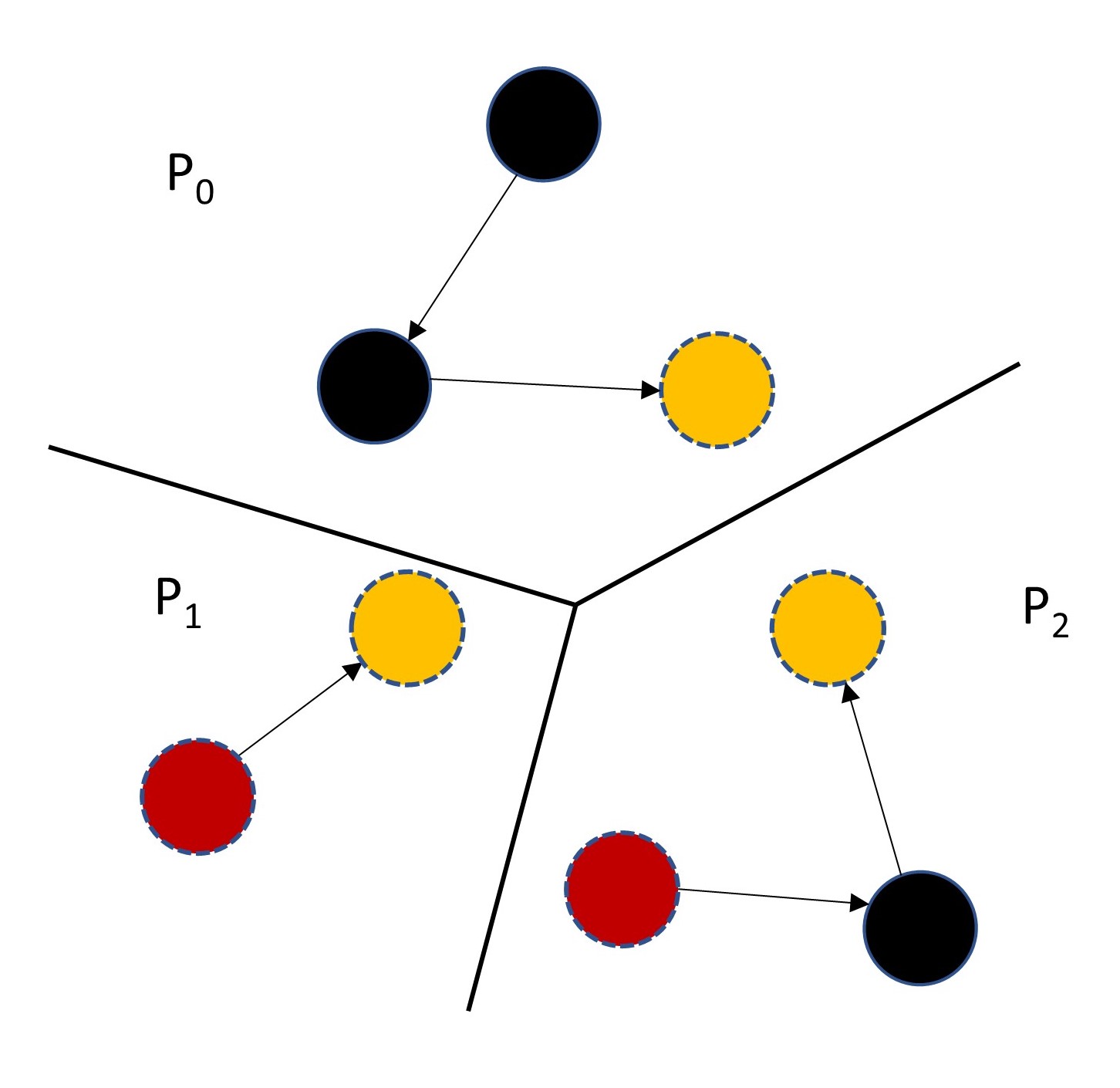}
\caption{Partitioned graph into three partitions}
\end{subfigure}
\caption{Example original and vertex-cut partitioned graph}
\label{fig-graph}
\end{figure}

\subsection{Partition Setup}

A partition contains two types of vertices: split-vertices and non-split vertices.
Split-vertices have their own copy of the feature vector. Thus, each vertex in a partition is associated with a feature vector and takes part in local aggregation.
All split-vertices communicate to receive feature vectors from their neighborhood and perform aggregation.
Vertices in each partition have global and local IDs. Global IDs are the vertex IDs from the original input graph. We assign consecutive local IDs to vertices starting from partition $0$ to partition $n-1$.
A global data structure {\tt vertex\_map} stores the local ID range of each partition. A vertex's local ID, along with the {\tt vertex\_map}, pinpoints its partition and location within the partition. 
For each split-vertex, we use an array to store the local IDs of all its clones.

\subsection{Delayed Remote Partial Aggregates}
Once the input graph is partitioned, we assign each partition to a socket for training the model. We implement the following three distributed algorithms with varying communication intensity during the aggregation operation. 
\begin{enumerate}
    \item {\tt 0c}: During training, at each layer, each partition first performs local aggregation. After local aggregation, each split-vertex has aggregated from its local partial neighborhood. \zeroc{} completely avoids communication by ignoring remote partial neighborhoods of  split-vertices.
    Due to no communication and its related pre- and post-processing computations, \zeroc{} is the fastest of all the three proposed algorithms and provides a performance roofline for scaling GraphSAGE on the given dataset across multiple \CPU{} sockets. It is also the most optimistic with respect to accuracy. 
    We evaluate its accuracy in section~\ref{sec-results}. 

    \item  {\tt cd-0}: During model training, at each layer, each partition perfoms local aggregation. All split-vertices communicate partial aggregates to their remote clones. Consequently, each vertex in each partition receives its complete neighborhood. \cdzero{} waits for communication to be completed before moving to the next step.
    Since each vertex receives its complete neighborhood, it is expected to produce the same accuracy as the single socket algorithm. \cdzero{} provides a lower-bound performance for scaling GraphSAGE on the given dataset across multiple \CPU{} sockets.
    
    \item {\tt cd-r}: Even after partitioning, the communication cost could be overwhelming (demonstrated in section~\ref{sec-results}). To further reduce communication volume, we apply a communication avoidance mechanism. In this algorithm, we overlap remote aggregate communication with local computation. The dependence between consecutive steps in an epoch leaves no scope for intra-epoch overlap. Consequently, we apply inter-epoch compute-communication overlap. Inspired by Hogwild~\cite{hogwild}, \cdr{} delays communicating partial aggregates among split-vertices; each split-vertex starts communication in epoch $i$ and asynchronously receives and processes aggregates in epoch $(i+r)$. 
    Communication can be further reduced by involving only a subset of split-vertices (through binning) in each epoch.
    We assess the impact on accuracy in section~\ref{sec-results} and demonstrate that accuracy is within $1$\% of the state-of-the-art baseline for each dataset. 
\end{enumerate}

The operation of distributed aggregation using the Delayed Remote Partial Aggregates (\DRPA{}) algorithm is described in Algorithm~\ref{algo:rpa}. 
A 1-level tree structure facilitates communication among the split-vertices in \cdzero{} and \cdr{}. 
For each original vertex $i \in v_s$, we create a tree $T_i$ in which we randomly assign one of its split-vertices as the root, while the rest of them become leaves. To synchronize aggregates across split-vertices, communication between leaves and the root occurs in two phases: (i) all leaves send their partial aggregates to the root (Line~\ref{async1}), (ii) the root receives and aggregates them and then communicates the final aggregates back to the leaves (Lines~\ref{async2}-\ref{async3}). 
Each partition performs a {\em pre-processing} and {\em post-processing} step for each partial aggregate communication. 
The pre-processing step involves local gather operation (Lines~\ref{pre1},~\ref{pre2}); 
it gathers features of split-vertices. Post-processing involves a local scatter-reduce operation (Lines~\ref{post1},~\ref{post2}). All received partial aggregates are scattered and reduced to corresponding vertices. 
Note that, only scatter operation is performed in the post-processing of root to leaf communication (Line~\ref{post2}).
\DRPA{} behaves like \cdzero{} when there is no delay for the communication i.e. $r=0$. When $r>0$, \DRPA{} functions as \cdr{}, performing delayed partial aggregations. Ignoring all the communication and associated pre- and post-processing in \DRPA{} produces the functionality of \zeroc{}. 

\begin{algorithm}[h]
\caption{Delayed Remote Partial Aggregates}\label{algo:rpa}
\begin{algorithmic}[1]
\small
\REQUIRE Graph partitions $G_p$ along with the vertices $V_p$ and features $f_{V_p}$
\REQUIRE $v_s$, a set of original vertices of $G_p$ which get split
\REQUIRE Tree $T_i$ for vertex $v_{s}[i], \forall i \in [0, |v_s|)$. 
\REQUIRE Root features $T_i.root \in f_{V_p}$ and leaf features $T_i.leaf  \in f_{V_p}$
\REQUIRE Delay parameter $r$
\STATE Allocate $G_p$ per socket $c$
\FOR{each $c$ in parallel}
\STATE $k \leftarrow |v_s|/r$
\FOR{i=$1\ to\ r$}
\STATE $S_i \leftarrow \{T_{i*k} \ldots T_{(i+1)*k}\}$  
\ENDFOR
\FOR{epoch $e$}
\STATE $f_v  \leftarrow \oplus({\tt copy}(f_u), f_v)$  \   $\forall u, v \in V_p$ 
\STATE $i \leftarrow e\%r$
\STATE $f_{v_{sl}} \leftarrow $ {\tt gather}($S_i[j].leaf,\ \forall j$)  \label{pre1} 
\STATE ${\tt async\_send}(f_{v_{sl}}$)  \label{async1}     
\IF{$e \ge r$}
\STATE $S_i[j].root \leftarrow {\tt async\_recv}(f_{v_{sl}}), \forall j$   \label{async2} 
\STATE $f_V \leftarrow {\tt scatter\_reduce}(S_i[j].root), \forall j$ \label{post1} 
\STATE $f_{v_{sr}} \leftarrow {\tt gather}(S_i[j].root), \forall j$  \label{pre2} 
\STATE ${\tt async\_send}(f_{v_{sr}}$)   \label{async3} 
\ENDIF
 \IF{$e \ge 2\times r$}
\STATE $S_i[j].leaf \leftarrow {\tt async\_recv}(f_{v_{sr}}), \forall j$  \label{async4} 
\STATE $f_V \leftarrow {\tt scatter}(S_i[j].leaf), \forall j$  \label{post2} 
\ENDIF
\ENDFOR
\ENDFOR
\end{algorithmic}
\end{algorithm}

%% file: results.tex
\section{Experimental Evaluation}
\label{sec-results}

\subsection{Experiment Setup}
\label{sec-results-exp-setup}

We perform our single-socket experiments on Intel Xeon $8280$ \CPU{} @$2.70$ {\tt GHz} with $28$ cores (single socket), equipped
with $98$ {\tt GB} of memory per socket; the theoretical peak bandwidth to DRAM on this machine is $128$ {\tt GB}/s. The machine runs CetnOS $7.6$.
For distributed memory runs, we use a cluster with $64$ Intel Xeon $9242$ \CPU{} @$2.30$ GHz with $48$ cores per socket in a dual-socket system. Each compute node is equipped with $384$ {\tt GB} memory, and the compute nodes are connected through Mellanox HDR interconnect with DragonFly topology. The machine runs CentOS $8$.
We use a single-socket machine with memory capacity of 1.5{\tt TB} to measure the single-socket runtime for OGBN-Papers dataset.

We use GCC v7.1.0 for compiling \DGL{} and the backend PyTorch neural network framework from their source codes.
We use a recent release of \DGL{}v0.5.3 to demonstrate the performance of our solutions and PyTorch v1.6.0 as the backend \DL{} framework for all our experiments. 
We use PyTorch Autograd profiler to profile the performance of the applications.

\subsubsection*{Datasets}
Table ~\ref{gnn-datasets} shows the details of the five datasets: AM, Reddit, OGBN-Products, OGBN-Papers, and Proteins~\cite{hipmcl, ucb}, used in our experiments. HipMCL~\cite{hipmcl} generated Proteins graph using isolate genomes from IMG platform. It performs sequence alignment among the collected sequences to generate a graph matrix. Entries in the graph matrix are generated based on sequence similarity scores. 
In the absence of vertex embeddings, we randomly generate features for the Proteins dataset. 
The AM dataset contains information about artifacts in the Amsterdam Museum~\cite{deBoer2012}. Each
artefact in the dataset is linked to other artifacts and details about its production, material, and content. It also has an artifact category, which serves as a prediction target. For this dataset, in the absence of vertex features, vertex ID is assigned as the feature.

\subsubsection*{Models and Parameters}
In the GraphSAGE model,
we use two graph convolutional layers for the Reddit dataset with $16$ hidden layer neurons. For the remaining datasets, we use three layers with the $256$ hidden layer neurons. In this paper, we employed \GCN{} aggregation operator where (i) $\oplus$ is {\em element-wise sum} and (ii) as a post-processing step, it adds the aggregated and original features of each vertex and normalizes that sum with respect to the in-degree of the vertex.
All the reported epoch run-times are averaged over $1$-$10$ epochs for \zeroc{} and \cdzero{} algorithms, while for \cdr{}  we average run-time for epochs $10$-$20$ due to the communication delay of $5$. 

\begin{table}[htb]
\caption{\GNN{} benchmark datasets. Edges are directed. Each original un-directed edge of Reddit, OGBN-Products, and Proteins is converted into two directed edges.}
\vspace{-10pt}
\begin{tabular}{l|c|c|c|c}
\hline
Datasets	     &  \#Vertex	    &  \#edge	  & \#feat &	\#class \\ \hline 
AM               & 881,680 & 5,668,682 & 1 & 11 \\
Reddit	         &  $\numprint{232965}$      &	$\numprint{114615892}$ &	$\numprint{602}$ &	$\numprint{41}$ \\
OGBN-Products    &	$\numprint{2449029}$     &	$\numprint{123718280}$ & $\numprint{100}$	&   $\numprint{47}$ \\
Proteins         &	$\numprint{8745542}$	    & $\numprint{1309240502}$  &	$\numprint{128}$ &	$\numprint{256}$ \\
OGBN-Papers  &	$\numprint{111059956}$	& $\numprint{1615685872}$  & $\numprint{128}$ &	$\numprint{172}$ \\
\hline
\end{tabular}
\label{gnn-datasets}
\end{table}

\subsubsection*{Implementation in \DGL{}}

All our optimized single code is written in {\tt C++} as part of \DGL{}'s backend. Our optimizations use  LIBXSMM library ~\cite{libxsmm}.

For distributed code, we use \texttt{torch.distributed} package of PyTorch coupled with Intel {\tt OneCCL}~\cite{torch-ccl} for efficient collective communication operations, with one MPI rank per socket;  two cores on each socket are dedicated to {\tt OneCCL}.
We use {\tt AlltoAll} collective for communicating the partial aggregates between the root and leaves in the 1-level tree. 
For parameter sync among the models, in each epoch, we use {\tt AllReduce} collective operation. All the distributed code is written in Python and {\tt C++} in \DGL{}. Our code is available at <url>.


\subsection{Single-Socket Performance}


\subsubsection*{\textbf{Application performance}}    

\begin{figure}[!ht]
\centering
\begin{minipage}{0.8\linewidth}
{
	\subfloat[GraphSAGE on Reddit]
	{
		\includegraphics[width=0.38\linewidth]{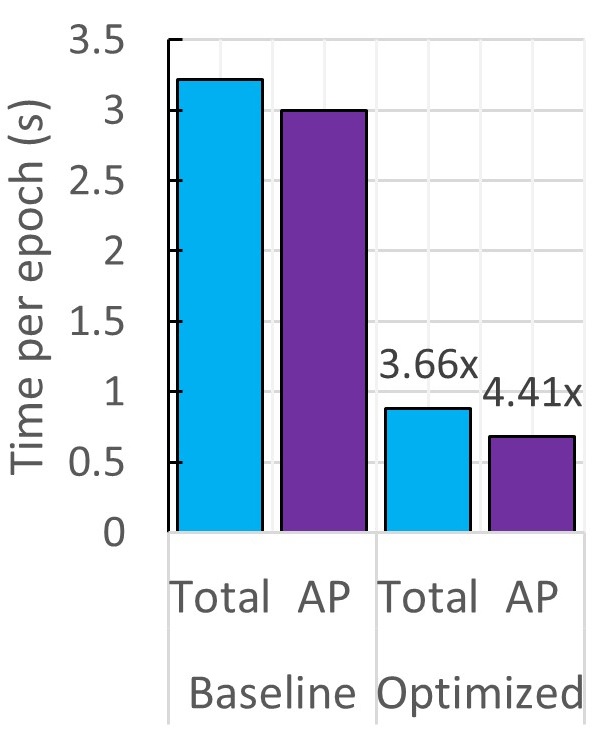}
		\label{fig:single-socket-reddit}
	}
	\hfill
	\subfloat[GraphSAGE on OGBN-Products]
	{
		\includegraphics[width=0.38\linewidth]{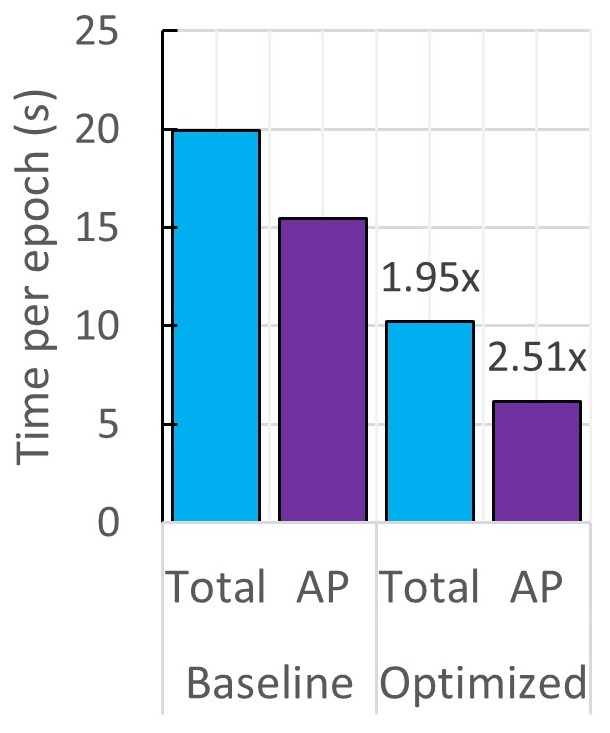}
		\label{fig:single-socket-ogbn-products}
	}
	\vfill
	\subfloat[GraphSAGE on Proteins]
	{
	    \includegraphics[width=0.38\linewidth]{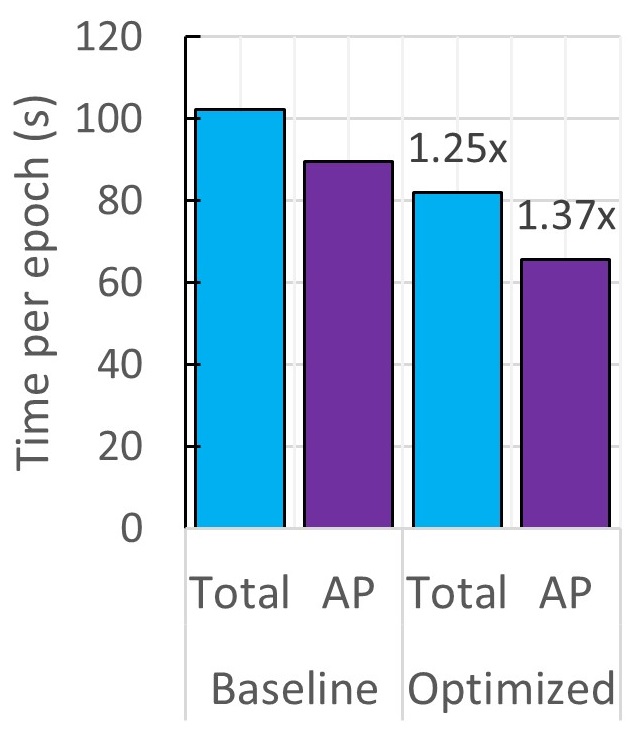}
        \label{fig:single-socket-proteins}
	}
	\hfill
	\subfloat[RGCN-hetero on AM]
	{
	    \includegraphics[width=0.38\linewidth]{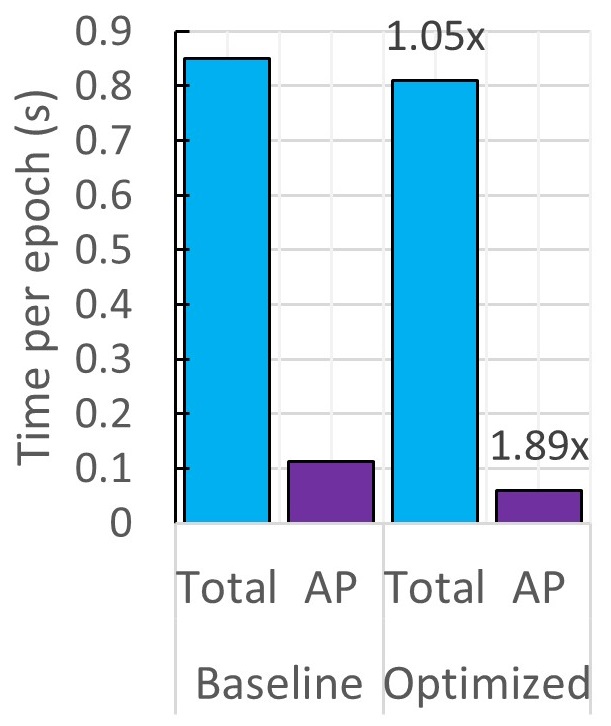}
        \label{fig:single-socket-am}
	}
}
\end{minipage}
\vspace{-10pt}
\caption{Comparison of runtime of entire training epoch (Total) and Aggregation Primitive (\AP{}) for baseline \DGL{} and our optimized version on the benchmark datasets. The labels on top of bars for optimized version show the speedup for Total time and \AP{} time, respectively.}
\label{fig:single-socket}
\end{figure}

Figure ~\ref{fig:single-socket} compares the per epoch training time and execution time of \AP{} for \DGL{} 0.5.3 with our optimized implementation for four of our workloads that fit on a single socket with $98$GB memory. It confirms that the runtime of many applications is dominated by \AP{}. Our optimizations achieve up to $4.41\times$ speedup for \AP{}, thereby, achieving up to $3.66\times$ speedup for end-to-end training time over \DGL{} 0.5.3. Optimization of \AP{} for shared memory systems was also presented in ~\cite{avancha2020deep}. Due to lack of availability of source code, we can only compare with the results presented in that paper. ~\cite{avancha2020deep} reports training time of $2.7$ secs/epoch for Reddit achieving a speedup of $7.7\times$ over \DGL{} 0.4.3, while we consume only $0.88$ secs/epoch. Compared to the \DGL{} 0.4.3, our current performance on Reddit is nearly $24\times$ faster. 

In the following, we explain the factors that affect performance gain of \AP{} kernel using the two benchmarks that achieve the highest speedup. Analysis of access to edge features is straightforward since they are accessed in streaming fashion. Therefore, we focus only on the vertex features by using $\otimes$ as copylhs. Without loss of generality, we use $\oplus$ as sum in our analysis.

\subsubsection*{\textbf{Effect of block size}}

\begin{table}[htbp]
\caption{Cache reuse achieved for the \AP{} kernel with respect to density of graph and the number of blocks ($n_B$). Density is defined as the number of non zero cells divided by total cells in the adjacency matrix. Ideal cache reuse is the average vertex degree of the graph -- 492 for Reddit and 50.5 for OGBN-Products.}
\vspace{-10pt}
\begin{tabular}{c|c|ccccccc}
\hline
 &  & \multicolumn{7}{c}{$n_B$}  \\
 \hline
Dataset & Density & 1 & 2 & 4 & 8 & 16 & 32 & 64 \\
\hline
Reddit & 0.002 & 3.1 & 4.3 & 7.3 & 16.1 & 27.0 & 16.7 & 9.6 \\
OGBN- & 0.00002 & 2.3 & 2.2 & 2.2 & 2.1 & 2.1 & 2.0 & 1.8 \\
Products & & & & & & & &  \\
\hline
\end{tabular}
\label{table:bw-analysis}
\end{table}

\begin{figure}[!ht]
\centering
\begin{minipage}{0.8\linewidth}
{
	\subfloat[Reddit]
	{
		\includegraphics[width=\linewidth]{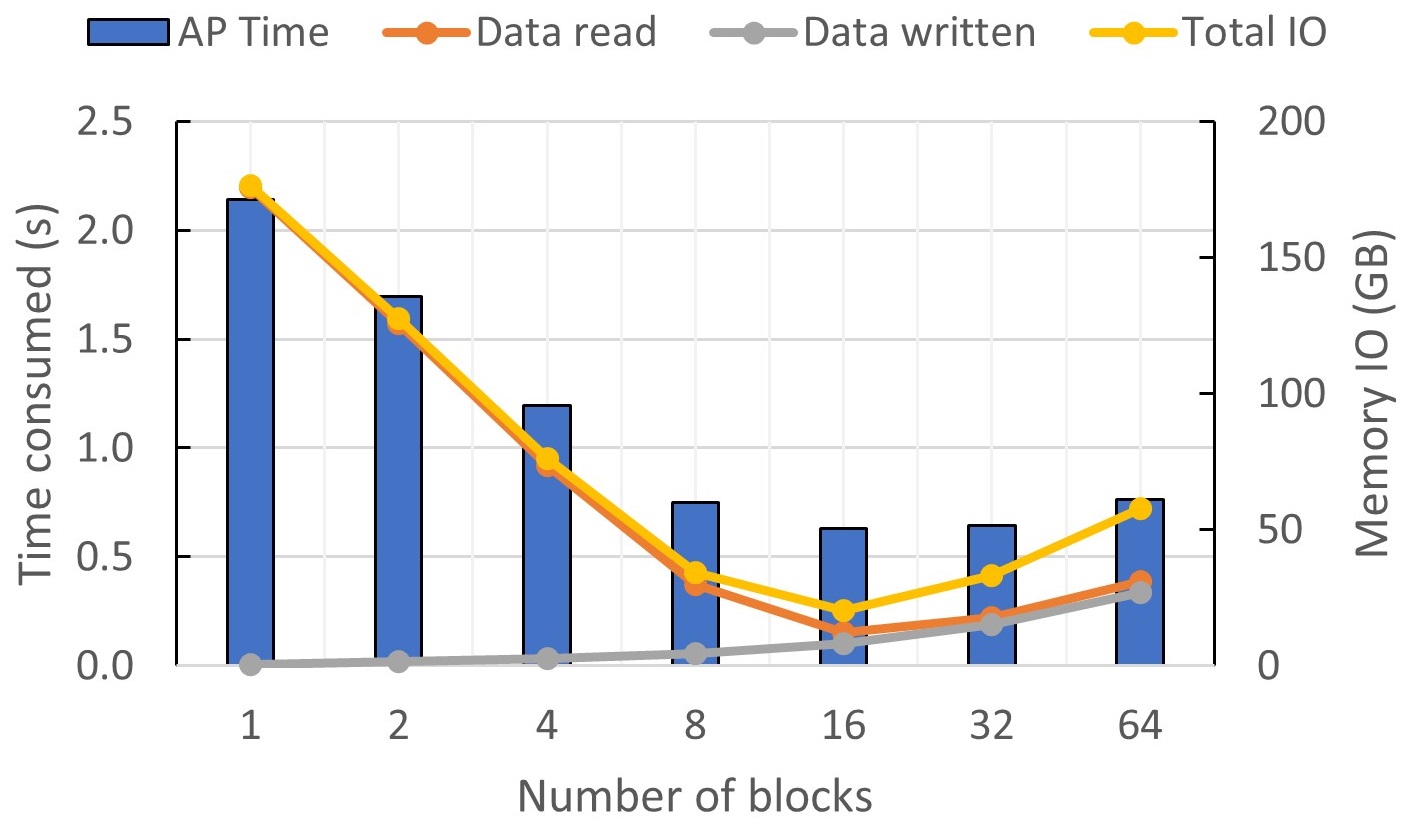}
		\label{fig:bw-reddit}
	}
	\vfill
	\subfloat[OGBN-Products]
	{
	    \includegraphics[width=\linewidth]{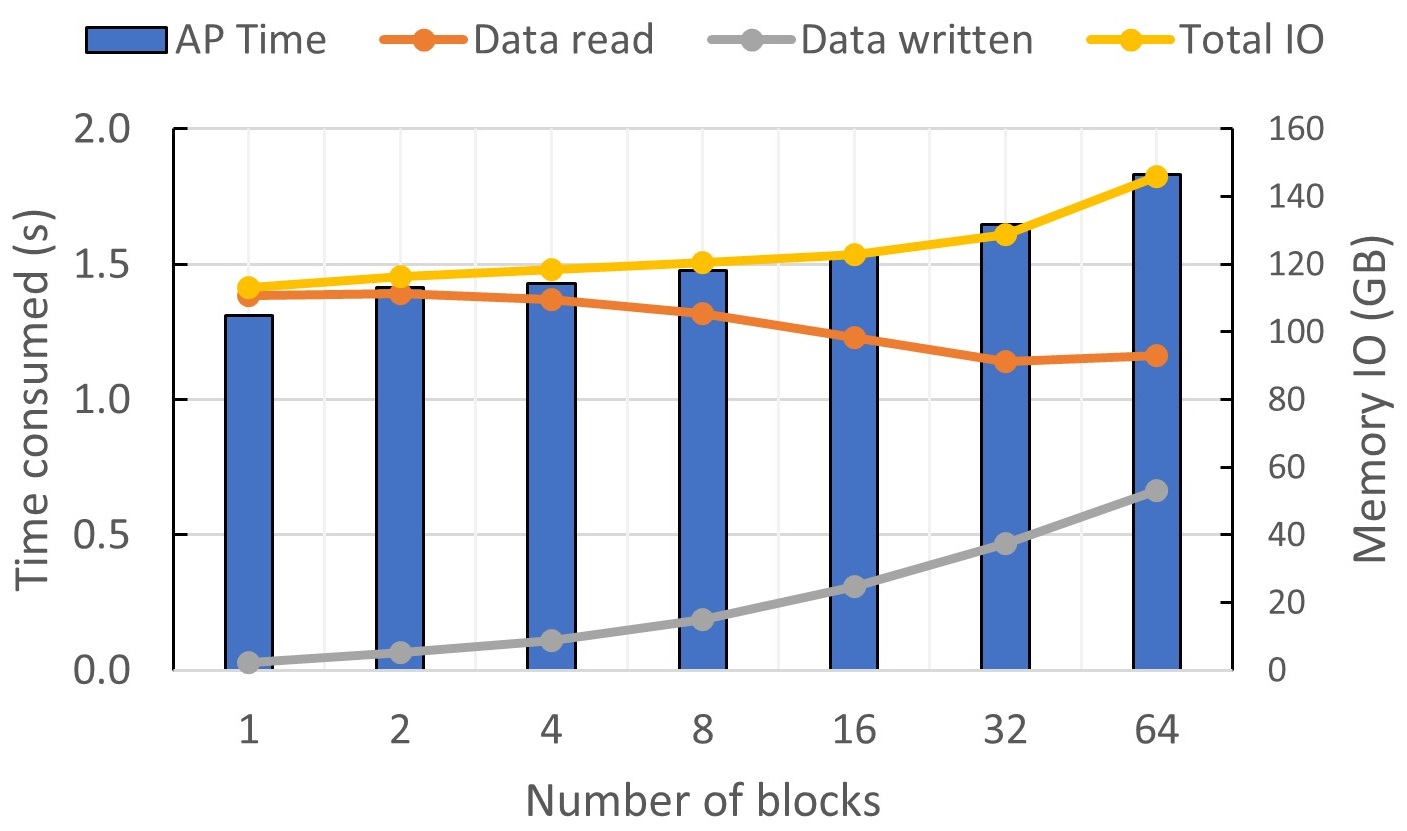}
        \label{fig:bw-ogbn-products}
	}
}
\end{minipage}
\vspace{-10pt}
\caption{Time consumed and amount of data read, written and total memory IO (data read + written) for \AP{} with respect to the number of blocks ($n_B$).}
\label{fig:bw-analysis}
\end{figure}

Table ~\ref{table:bw-analysis} and Figure ~\ref{fig:bw-analysis} illustrate the effect of block size ($B$), and hence, the number of blocks ($n_B$). In the ideal scenario, every feature vector ($f_\mathcal{V}[u]$) would be loaded only once from memory and would be used to update feature vectors in $f_\mathcal{O}$ of all its neighbors. Similarly, every feature vector in $f_\mathcal{O}$ would also be written to memory only once. Therefore, maximum average reuse of the feature vectors in $f_\mathcal{V}$ and $f_\mathcal{O}$ is the average vertex degree of the graph. While ideal reuse is not possible, our blocking method tries to maximize reuse. Clearly, the sparser the graph, the less likely it is to reuse the feature vectors. When we use just one block, the blocks of $f_\mathcal{V}$ start getting thrashed out of cache after processing a few rows of the adjacency matrix, thus, preventing reuse. As the blocks get smaller, chances of blocks of $f_\mathcal{V}$ staying in the cache increase, thereby, increasing reuse in $f_\mathcal{V}$. At the same time, we need more passes of $f_\mathcal{O}$ decreasing its reuse. This gets reflected in data read from and written to memory as clearly seen in Figure ~\ref{fig:bw-analysis}. The best performance is at the sweet spot where the sum of data read and written is the smallest. For denser graphs (like Reddit), the sweet spot is more to the right compared to sparser graphs (like OGBN-Products).

\subsubsection*{\textbf{Breakup of speedup with respect to optimizations}}

\begin{figure}[!ht]
\centering
\includegraphics[width=0.8\linewidth]{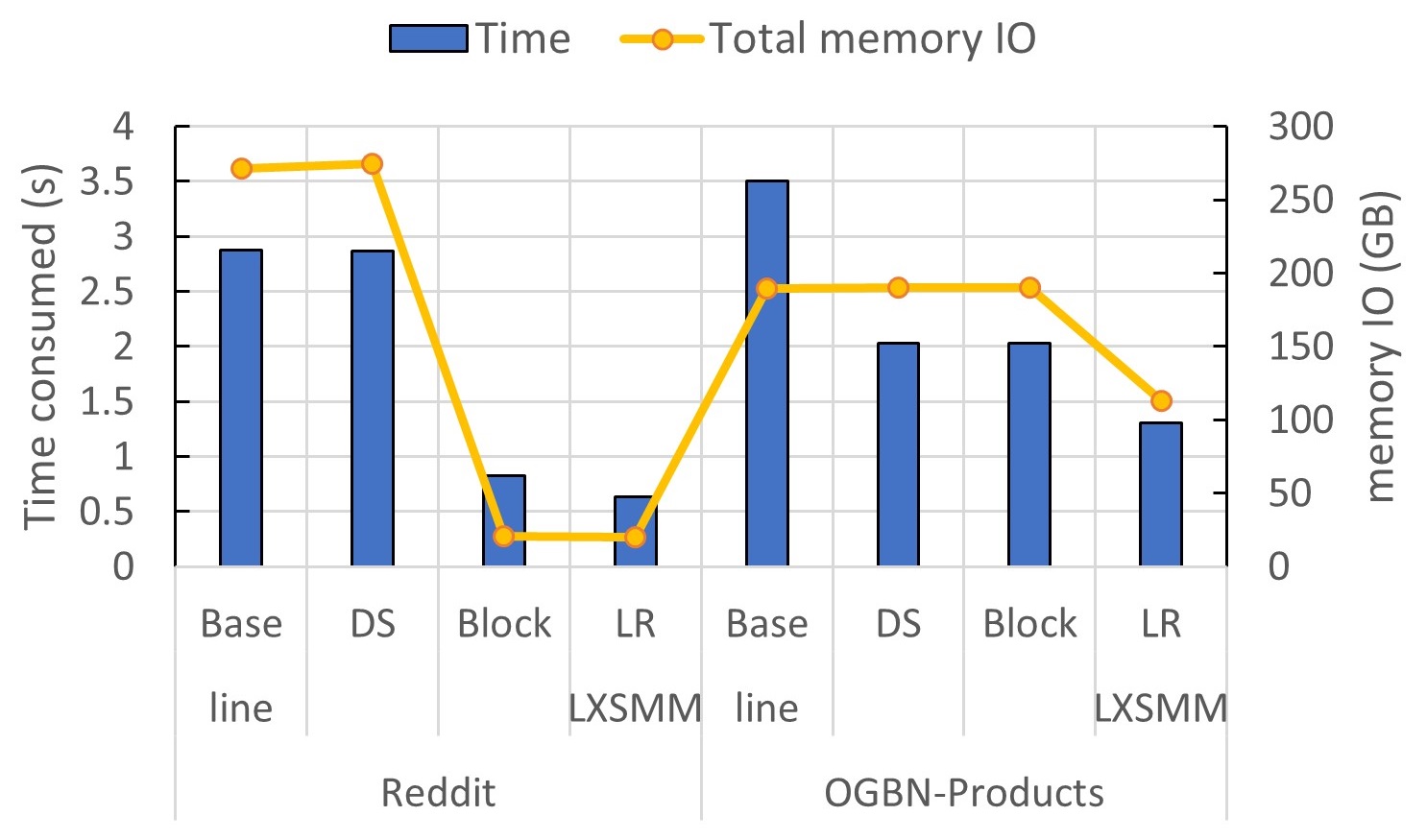}
\vspace{-10pt}
\caption{Effect of Dynamic Scheduling (DS), Blocking (Block) and Loop Reordering with LIBXSMM (LR LXMM) on the total memory IO and execution time of \AP{} for Reddit and OGBN-Products.}
\label{fig:opt-benefit}
\end{figure}

Figure ~\ref{fig:opt-benefit} shows the breakup. There is a clear correlation between memory IO required and execution time. Dynamic scheduling has no effect for Reddit but plays a major role in improving performance for OGBN-Products. On the other hand, blocking has a massive impact for Reddit but has no effect for OGBN-Products where we end up using only one block. Loop reordering and JITing with LIBXSMM improves the performance in both cases.

\subsection{Distributed Algorithm Performance}

\begin{table}[htb]
\caption{Average replication factor due to vertex-cut based graph partitioning using Libra}
\vspace{-10pt}
\begin{tabular}{l|c|c|c|c|c|c|c}
\hline
Datasets/	&\multirow{2}{*}{2}		&\multirow{2}{*}{4}		&\multirow{2}{*}{8}		&\multirow{2}{*}{16}		&\multirow{2}{*}{32}	&\multirow{2}{*}{64}		&\multirow{2}{*}{128} \\ 
\#Partitions & & & & & & & \\ \hline
Reddit & 1.75 & 2.94 & 4.66 & 6.93 & - & - & - \\ 
OGBN-Products & 1.49 & 2.16 & 2.98 & 3.90 & 4.85 &5.74 & - \\ 
Proteins & 1.33 & 1.65 & 1.91 & 2.11 & 2.27 & 2.37& - \\ 
OGBN-Papers & - & - & - & - & 4.63 & 5.63 & 6.62 \\ 
\hline
\end{tabular}
\label{tab:graph-partition}
\end{table}

\begin{figure*}[!ht]
\begin{minipage}{0.48\textwidth}
    \caption*{Reddit}
    \centering
    \includegraphics[width=\textwidth]{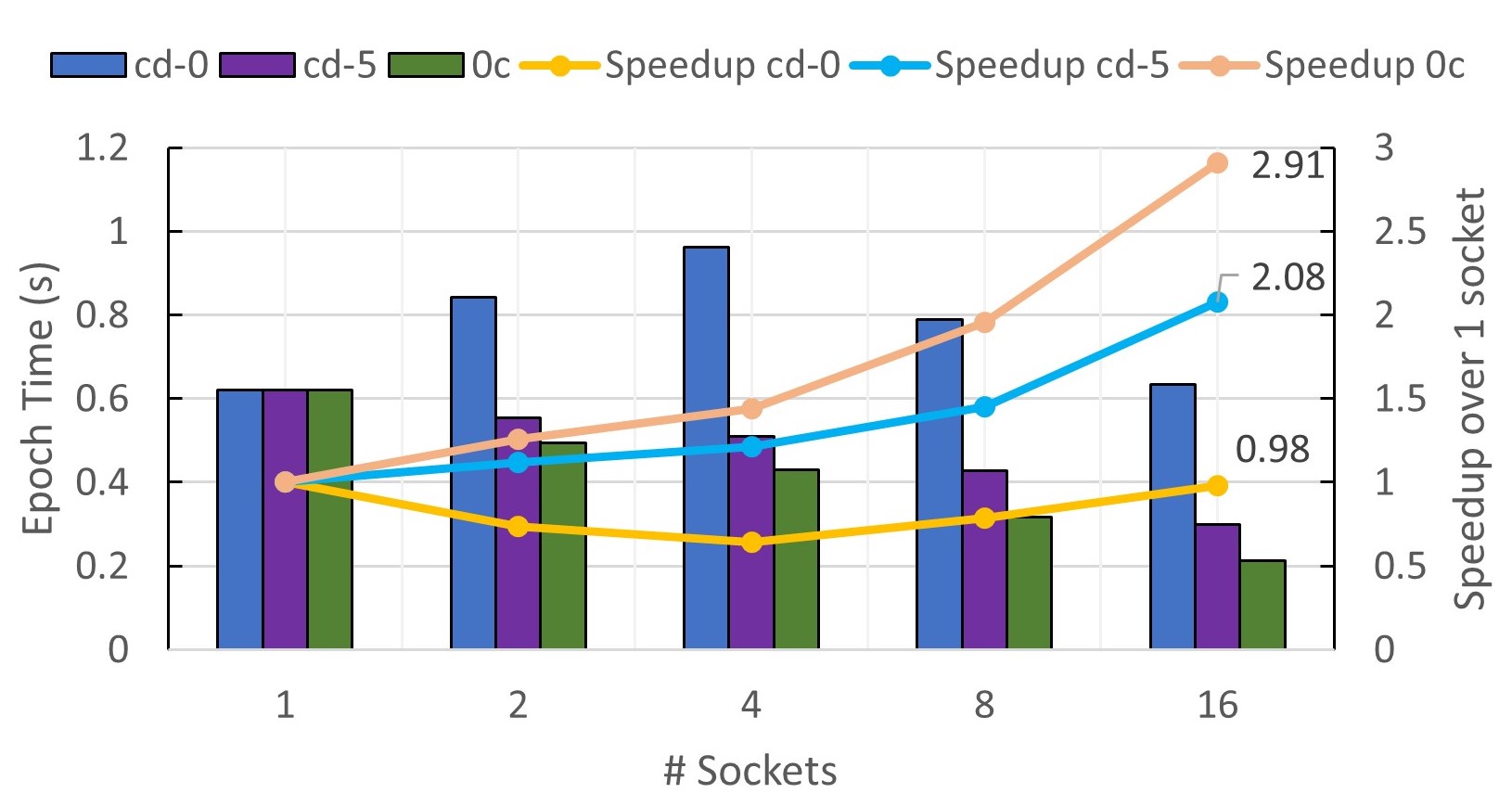}
    \label{fig:reddit1}
\end{minipage}
\begin{minipage}{0.48\textwidth}
    \caption*{OGBN-Products}
    \centering
    \includegraphics[width=\textwidth]{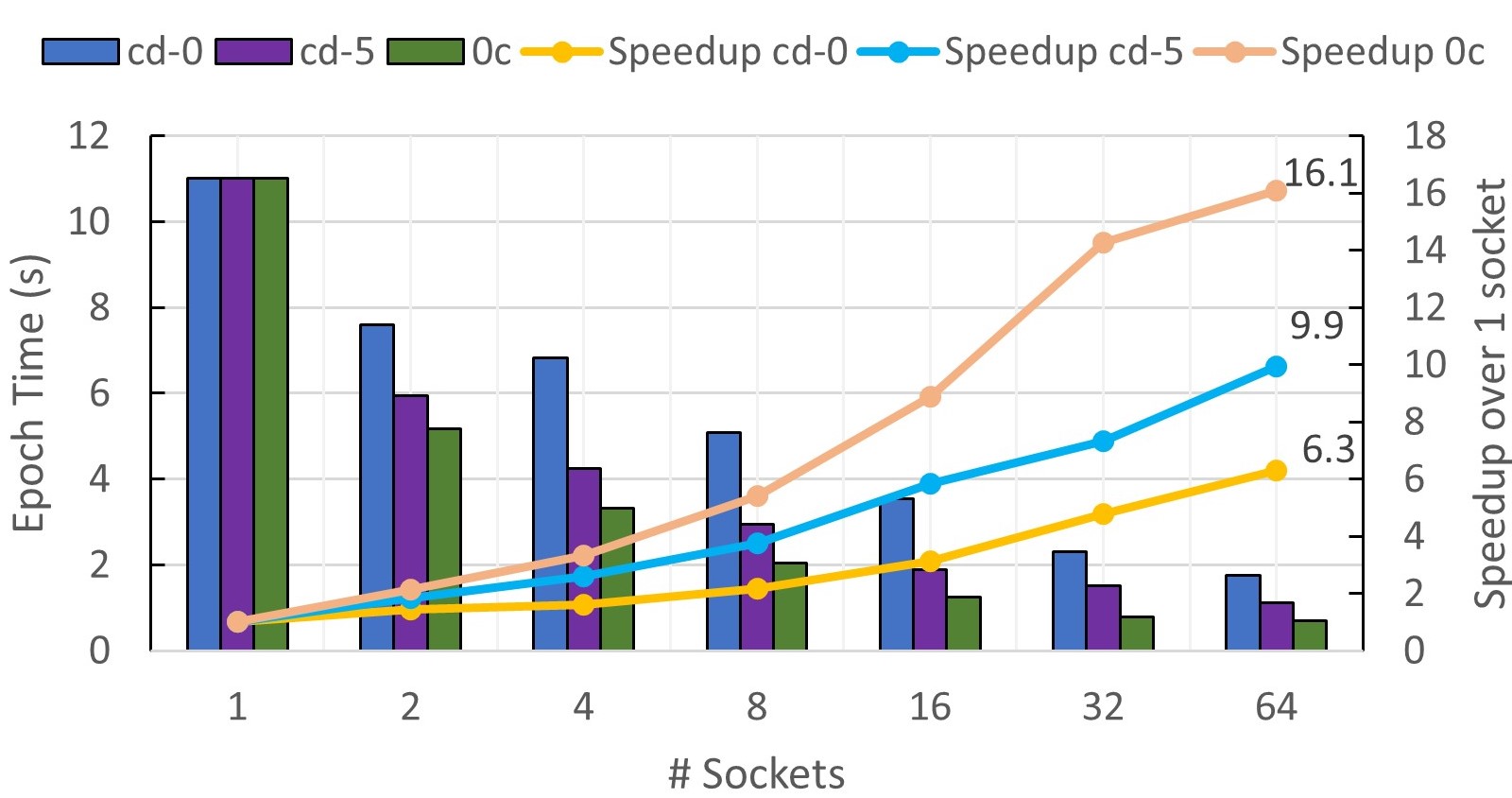}
    \label{fig:products1}
\end{minipage}
\begin{minipage}{0.48\textwidth}
    \vspace{-20pt}
    \caption*{Proteins}
    \centering
    \includegraphics[width=\textwidth]{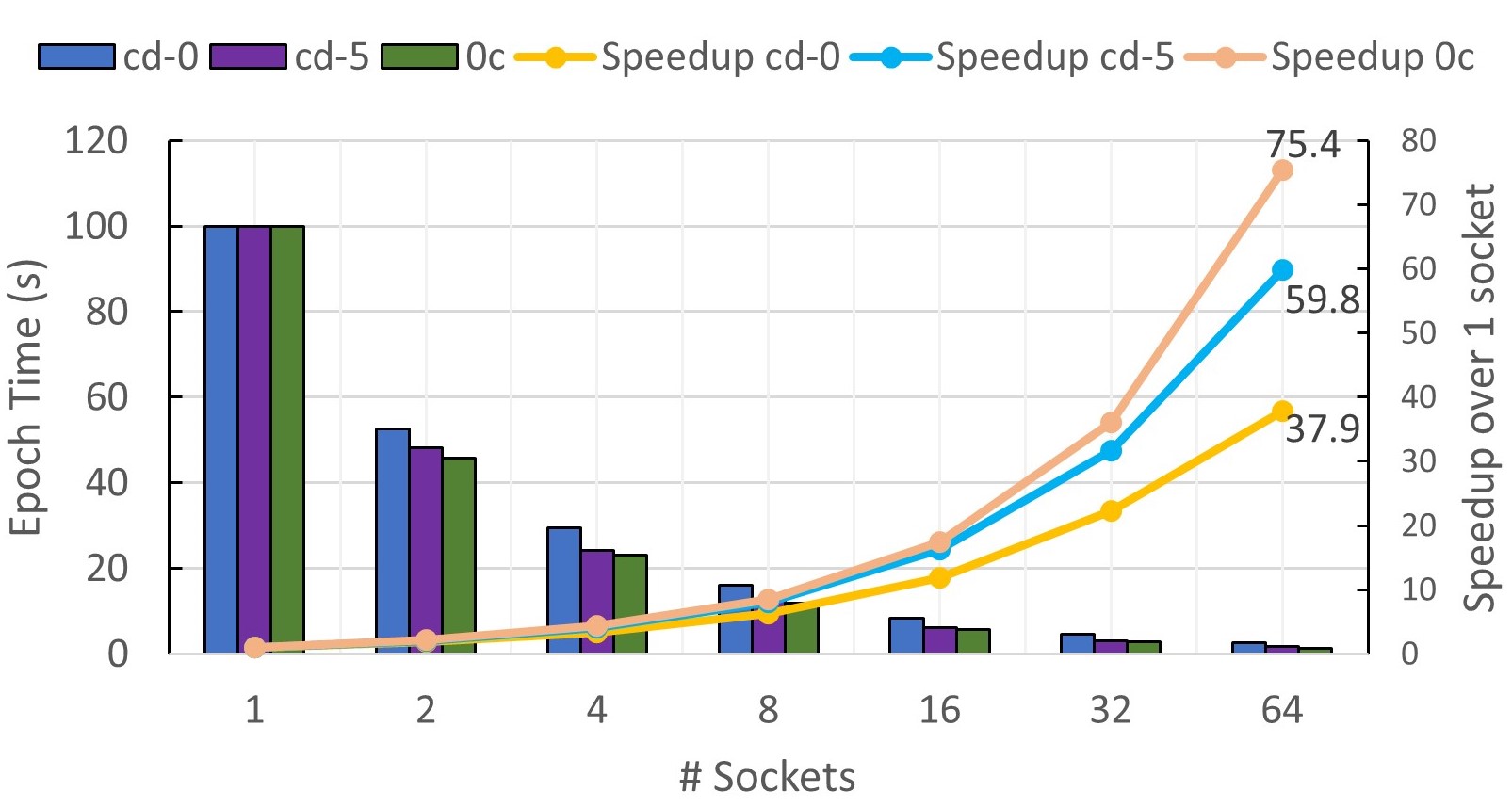}
    \label{fig:proteins1}
\end{minipage}
\begin{minipage}{0.48\textwidth}  
    \vspace{-20pt}
    \caption*{OGBN-Papers}  
    \centering
    \includegraphics[width=\textwidth]{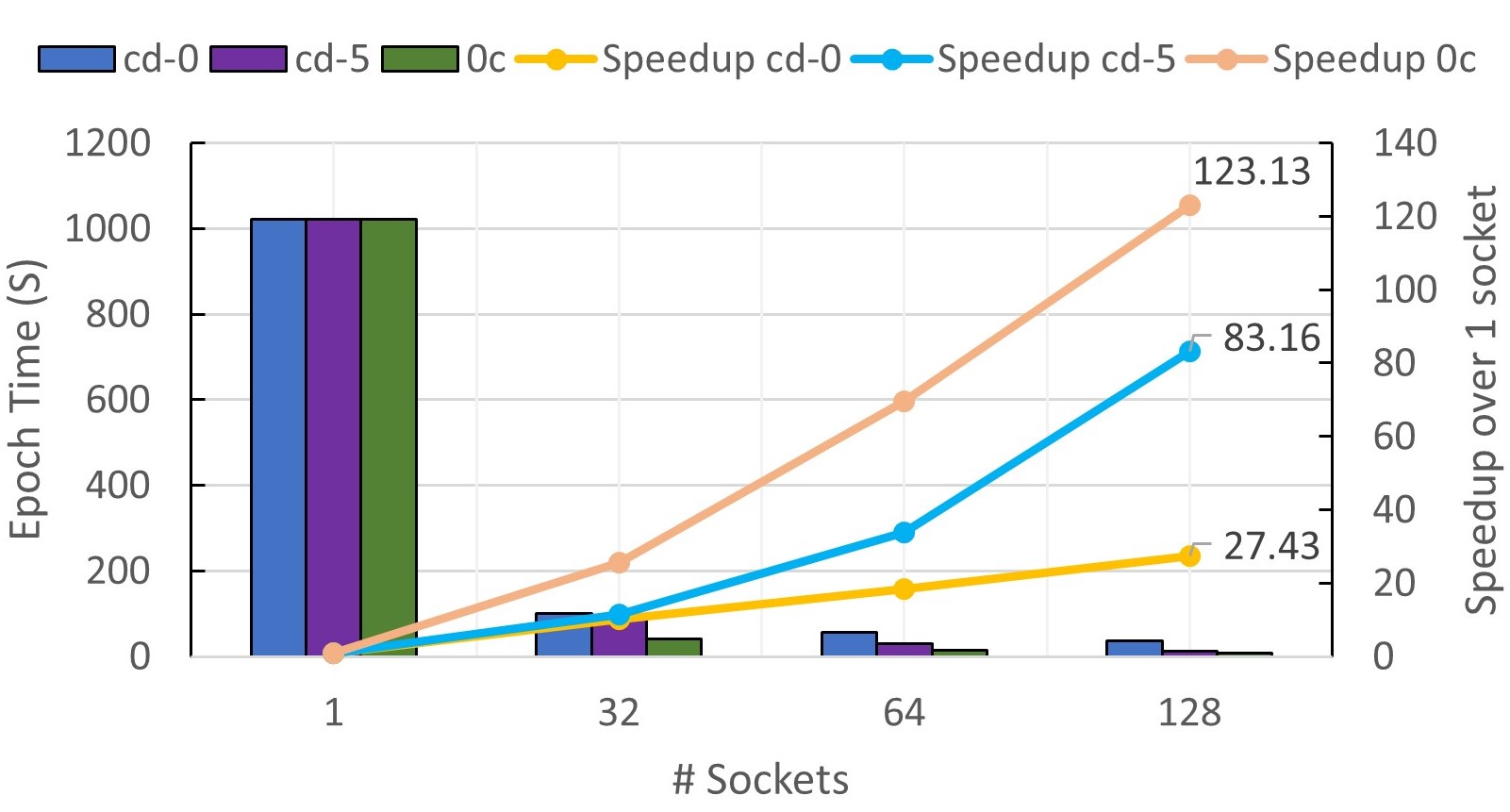}
    \label{fig:papers1}
\end{minipage}%
\vspace{-20pt}
\caption{Runtime performance and speedup of three distributed algorithms of \DistGNN{} on the benchmark datasets. 
}
\label{fig-scaling}
\end{figure*}

\subsubsection*{\textbf{Graph Partition}}
As seen in the previous sections, data movement during aggregation is directly related to the number of edges, and hence the run-time. Thus, we use the simple criteria of equal edge allocation among the partitions as the load balancing mechanism.
Libra, despite having no hard constraints on maintaining an equal distribution of edges to the partitions, produces highly balanced partitions in terms of the number of edges.

Table~\ref{tab:graph-partition} shows the average vertex replication for a different number of partitions produced by Libra. Reddit, the densest of the benchmark datasets, results in relatively more split-vertices during partitioning than all other datasets.  Proteins results in a significantly smaller replication factor. It exhibits natural clusters of protein families (sequence homology), thus leading itself to high-quality partitioning.
OGBN-Products and OGBN-Papers, with the least average vertex degree, have similar replication factors.
A lower replication factor is desirable.
On single socket, a higher replication factor leads to a sparser partitioned graph which results in more pressure on the memory bandwidth, whereas in a distributed setting, an increase in replication factor with the number of partitions leads to more communication and hampers the scalability of the solution.

\subsubsection*{\textbf{Scaling}}
In all our experiments, we run \cdr{} algorithm with delay of $r=5$ epochs.
Figure~\ref{fig-scaling} shows the per epoch time and speed-up of our solutions with increasing socket-count.
Owing to the high replication factor in Reddit partitioning, the decrease in partition size from $2$ to $16$ partitions is highly sub-linear. This directly leads to a sub-linear decrease in local and remote aggregation time. 
For $16$ sockets, we observe $0.98\times$, $2.08\times$, and $2.91\times$ speed-up using \cdzero{}, \cdf{}, and \zeroc{}, respectively compared to the optimized \DGL{} single-socket performance.  
In contrast to Reddit, the Proteins dataset exhibits nearly linear decrease in partition size from $2$ to $64$ partitions. For $64$ sockets, we observe $37.9\times$, $59.8\times$, and $75.4\times$ speed-up using \cdzero{}, \cdf{}, and \zeroc{}, respectively compared to the optimized \DGL{} single-socket performance. Due to the usage of memory from multiple Non-Uniform Memory Access ({\tt NUMA}) domains, the single-socket run is slower, leading to a super-linear speedup for \zeroc{}.

The quality of partitions for OGBN-Products, as reflected in its replication factor, is in-between those of Reddit and Proteins datasets. OGBN-Papers has slightly better quality partitions compared to OGBN-Products. For OGBN-Products, for $64$ sockets, we observe $6.3\times$, $9.9\times$, and $16.1\times$ speed-up using \cdzero{}, \cdf{}, and \zeroc{}, respectively compared to the optimized \DGL{} single-socket performance. 
For OGBN-Papers, for $128$ sockets, we observe $27.43\times$, $83.16\times$, and $123.13\times$ speed-up using \cdzero{}, \cdf{}, and \zeroc{}, respectively compared to the optimized \DGL{} single-socket performance. Here, due to very high memory requirements ($1.4${\tt TB}), a single-socket run uses memory from multiple \NUMA{} domains and thus runs slower. Similarly, $32$ and $64$ socket runs also use memory from multiple \NUMA{} domains and run slower. Owing to high memory demands, we could not run OGBN-Papers dataset below $32$ sockets.
Note that for multi-socket runs, two cores per socket are reserved for {\tt OneCCL} library.

\begin{figure}[ht]
    \begin{minipage}{0.49\linewidth}
    \caption*{Reddit}
    \centering
    \includegraphics[width=\linewidth]{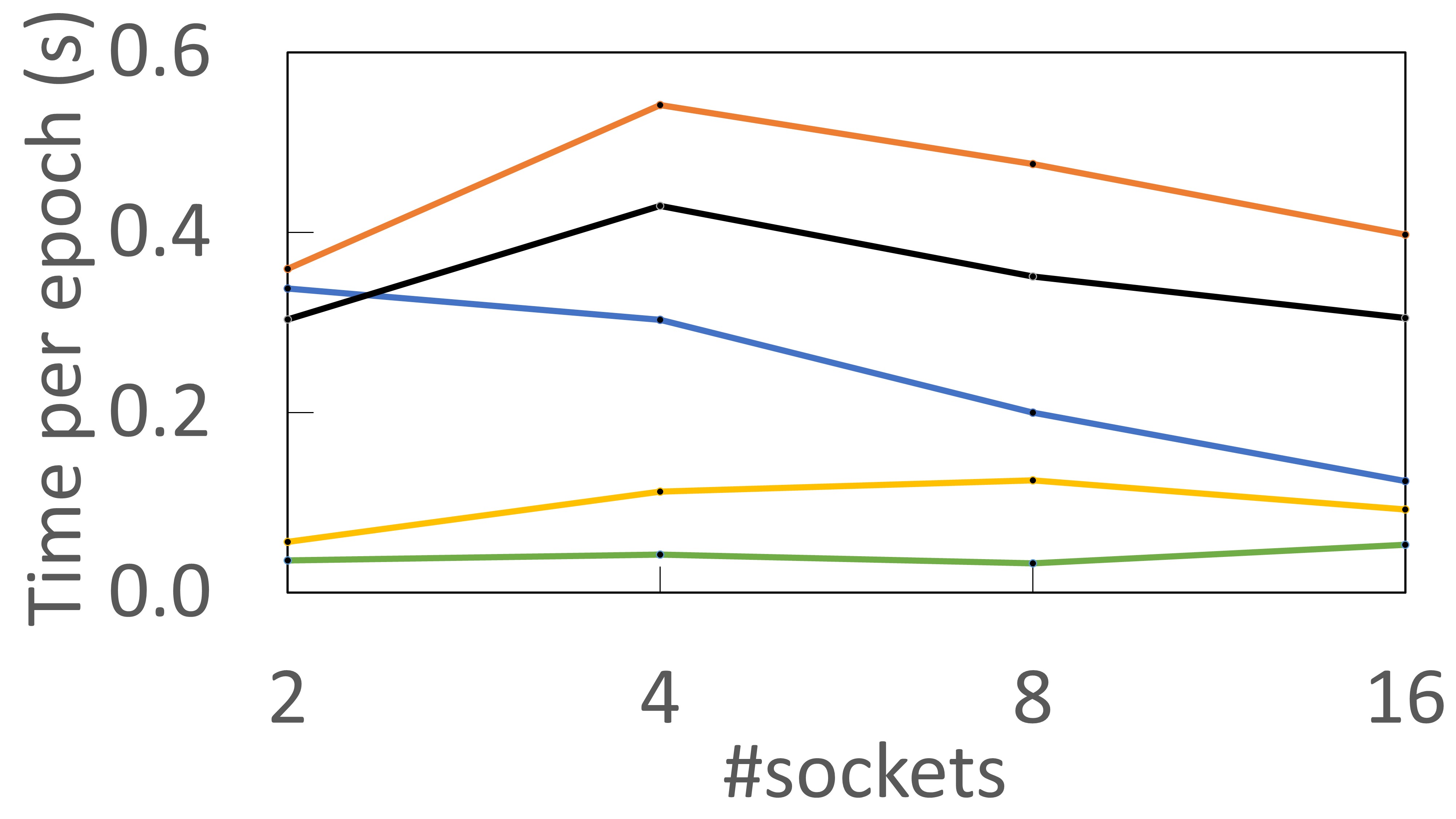}
    \label{fig:reddit2}
    \end{minipage}  
    \begin{minipage}{0.49\linewidth}
    \caption*{OGBN-Products}
    \centering
    \includegraphics[width=\linewidth]{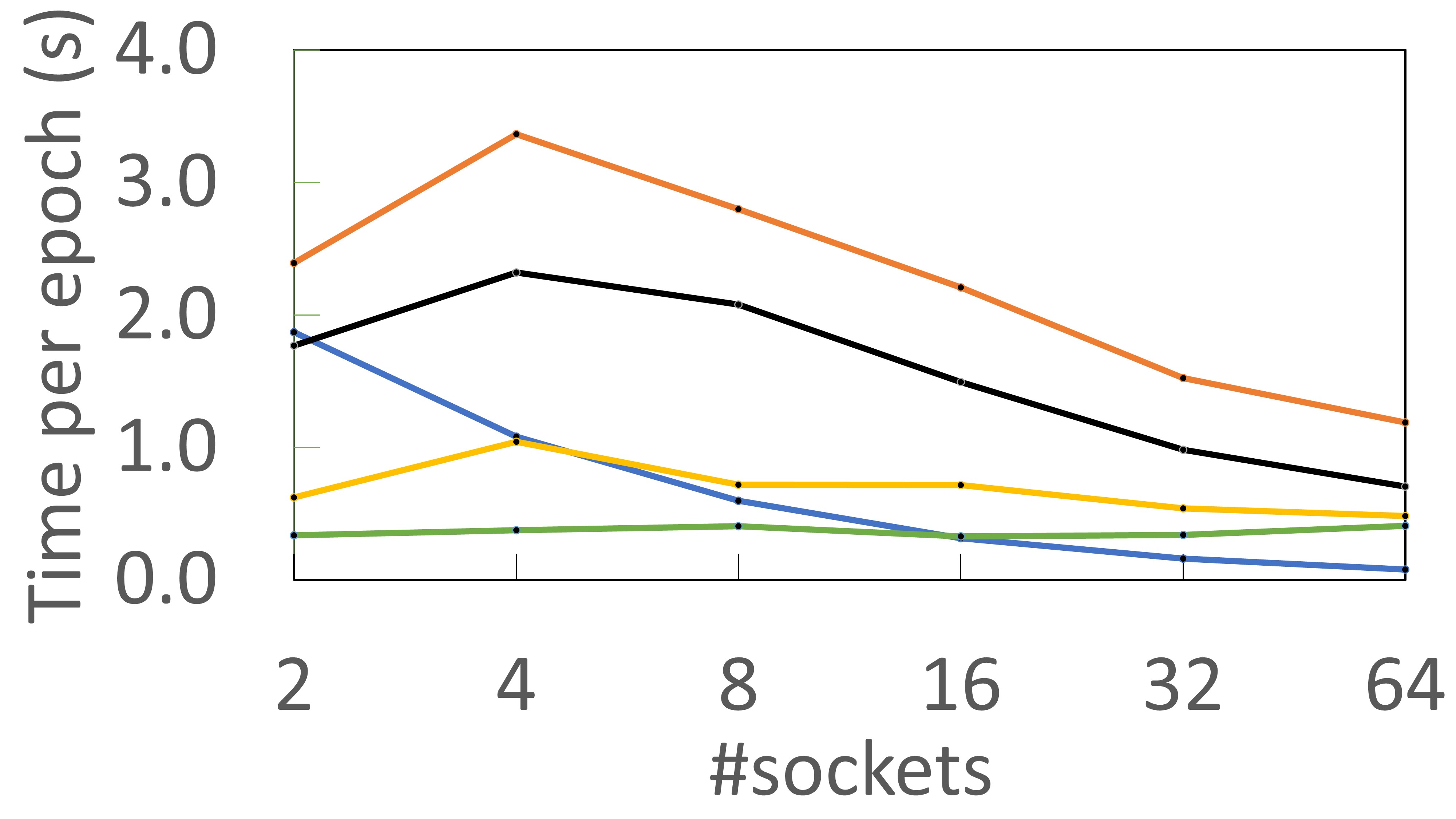}
    \label{fig:products2}
    \end{minipage}  
    \begin{minipage}{0.49\linewidth}
    \caption*{Proteins}
    \centering
    \includegraphics[width=\linewidth]{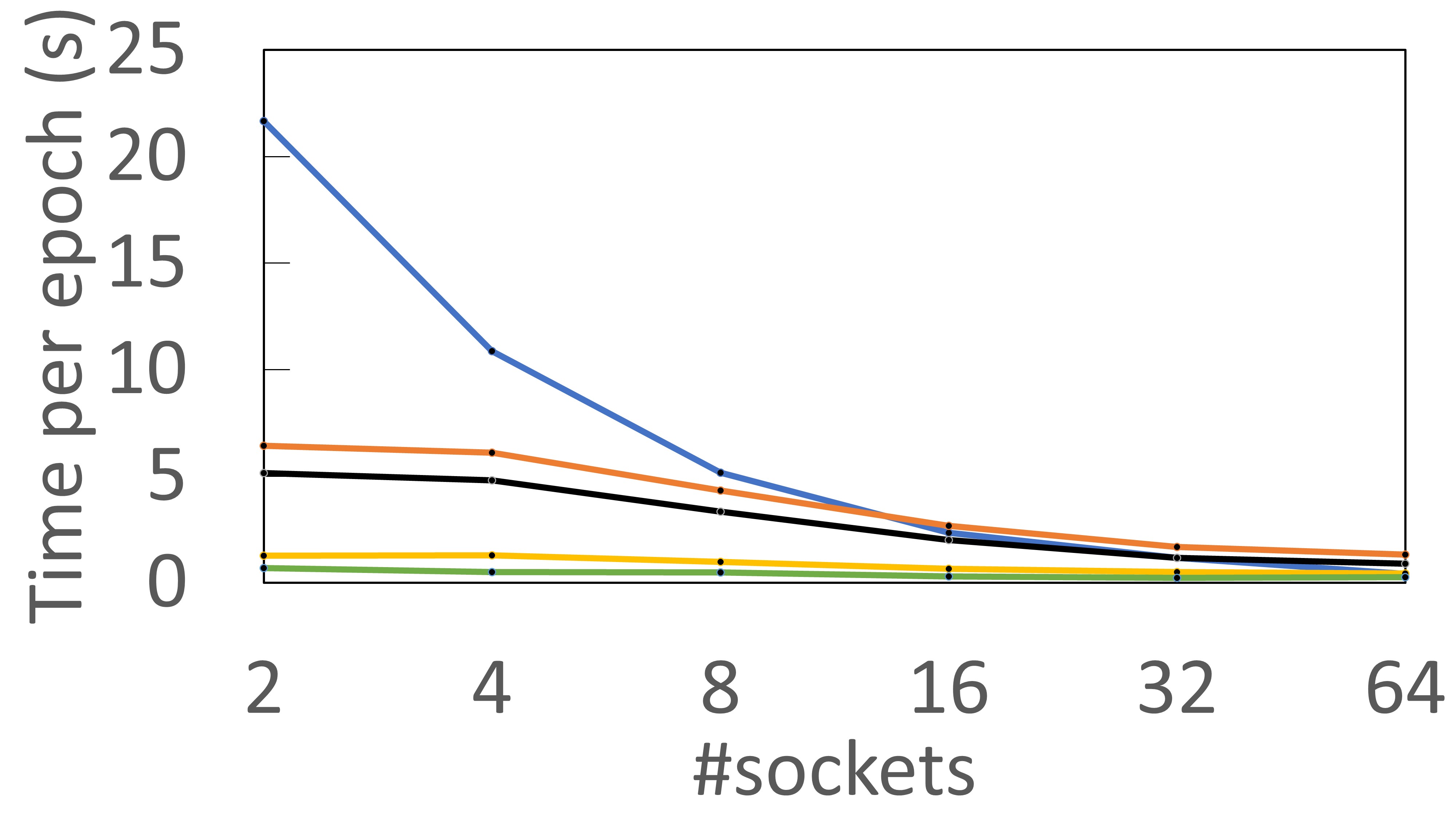}
    \label{fig:proteins2}
    \end{minipage}  
    \begin{minipage}{0.49\linewidth}
    \caption*{OGBN-Papers}
    \centering
    \includegraphics[width=\linewidth]{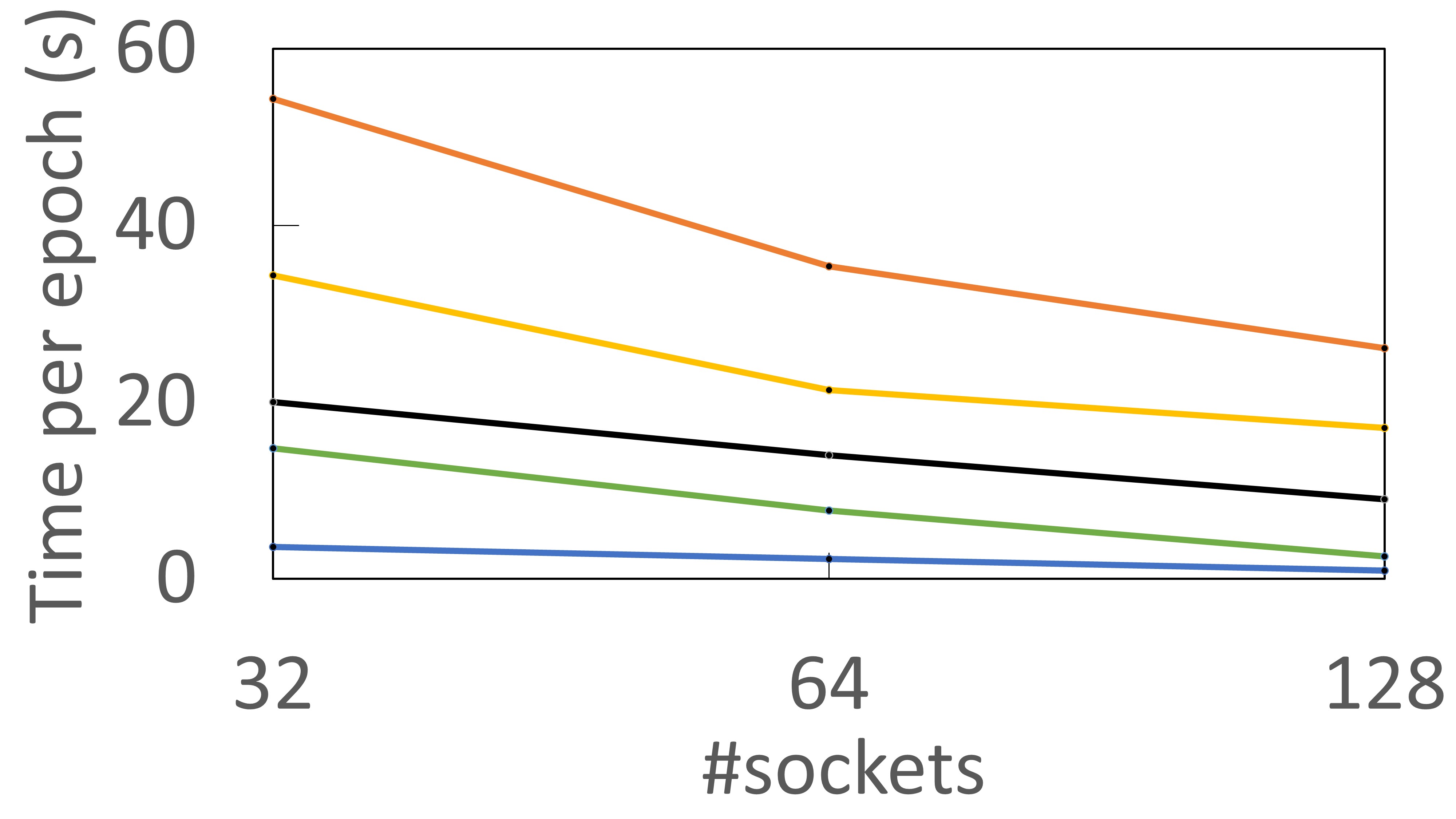}
    \label{fig:papers2}
    \end{minipage} 
    \begin{minipage}{\linewidth}
    \centering
    \includegraphics[width=\linewidth]{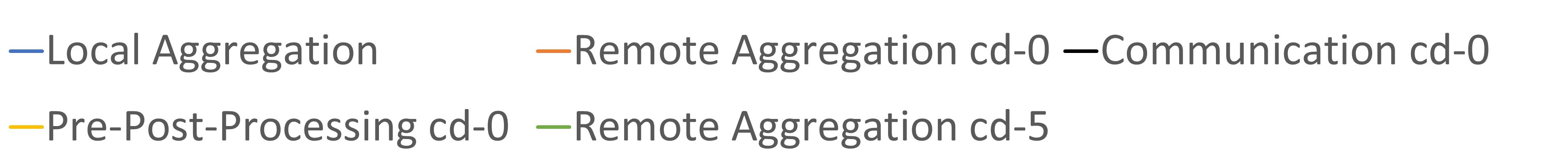}
    \end{minipage} %
    \vspace{-10pt}
    \caption{Forward pass scaling performance of local and remote aggregate operation of {\tt cd-0}, {\tt cd-5}, and \zeroc{} on benchmark datasets. The remote aggregation involve pre- and post-processing time for communication. Due to the absence of communication in \zeroc{}, its time is same as local aggregation time. 
    }
    \label{fig-component-scaling}
\end{figure}

As described in Section~\ref{sec-distgnn-distributed}, 
the aggregation step involves local and remote aggregation sub-steps in \cdzero{} and \cdr{}, whereas for \zeroc{}, it comprises of only the former sub-step. 
Remote aggregation not only involves actual communication but also pre- and post-processing.
Local aggregation does not involve communication, therefore local aggregation time (\LAT{})
remains the same across \cdzero{}, \cdr{}, and \zeroc{}. Figure~\ref{fig-component-scaling} shows that in the forward pass, \LAT{} scales linearly with the number of sockets, except for Reddit.
It also shows that remote aggregation time (\RAT{}) scales poorly with the increase in the number of sockets; this is solely an artifact of the replication factor and number of split-vertices per original vertex. For \cdf{}, we observe that a negligible amount of time is spent in waiting for asynchronous overlapped communication; thus, \RAT{} is purely composed of pre- and post-processing times. For Reddit and Proteins, local aggregation dominates remote aggregation. For OGBN-Products, remote aggregation consumes a significant portion of the overall execution time beyond $16$ sockets, reducing \AP{} scalability. For OGBN-Papers, \RAT{} is always higher than \LAT{}, due to much higher pre- and post-processing cost compared to other datasets. 
Due to high communication volume in \cdzero{}, \RAT{} is higher than \LAT{} for all datasets, except Proteins. 

\subsubsection*{\textbf{Accuracy}}
The GraphSAGE model reports single-socket test accuracy of $93.40\%$ and $77.63\%$ for Reddit and OGBN-Products respectively.
We evaluate the test accuracy of all the three distributed algorithms on a number of partitions (Table ~\ref{tab:accuracy}). We use the delay factor $r=5$ for our \cdr{} experiments. For both the datasets, all the distributed algorithms report the accuracy within $1\%$ of the best accuracy. For $8$ and $16$ sockets using Reddit dataset, we observe that accuracy recovers with increased training time from $200$ to $300$ epochs. Algorithms \cdf{} and \zeroc{}, for some cases, on both the datasets, report accuracy better than single-socket accuracy. One possible reason for this increase is clustering due to partitioning. In all the experiments with \cdr{}, we do not see any discernible improvements in accuracy with values of $r < 5$, while large values of $r$ (e.g., $r=10$)  degraded the accuracy due to increasingly stale feature aggregates. 

Due to the absence of the features and labeled data for the Proteins dataset, we were unable to validate training accuracy with test data. 
For OGBN-papers, we use GraphSAGE as the primary \GNN{} model  – same as the \DistDGL{}. We show how to scale OGB-papers across 128 sockets. While the accuracy of vanilla GraphSAGE (Table~\ref{tab:accuracy}) for this dataset is significantly lower than that reported in OGB leaderboard~\cite{ogbn-leadership}, there are techniques (such as~\cite{shi2020masked}) to bridge the accuracy gap, which we plan to explore.
Table~\ref{tab:accuracy} also reports the test accuracy attained by our distributed algorithms. Further experiments with hyperparameters tuning are required for OGBN-Papers dataset.     

\begin{table}[htb]
    \centering
    \caption{Test accuracy of single socket and distributed algorithms using Reddit, OGBN-Products, OGBN-Papers dataset, with corresponding learning rate ($lr$) and number of epochs. We set weight decay, $wd=5e^{-4}$ for all the experiments.}
    \vspace{-10pt}
    \begin{tabular}{p{20pt}|p{15pt}r|p{15pt}r|p{15pt}r|c}
    \hline
    \multicolumn{8}{c}{Reddit} \\ \hline   
    &\multicolumn{2}{c|}{\cdzero{}}	&		\multicolumn{2}{c|}{\cdf{}}	&		\multicolumn{2}{c|}{\zeroc{}} & \\ \hline		
    \#soc- kets	&Acc.(\%)	&$lr$	&	Acc.(\%)&	$lr$&	Acc.(\%)&	$lr$&	\#epochs \\ \hline
    1	&93.40	&0.01		&93.40	&0.01		&93.40	&0.01	&200  \\
    2	&93.70	&0.028		&93.59	&0.028		&93.58	&0.028	&200  \\
    4	&93.44	&0.028		&93.25	&0.028		&93.39	&0.028	&200  \\
    8	&93.14	&0.028		&93.33	&0.08		&93.14	&0.07	&300  \\
    16	&92.86	&0.028		&92.62	&0.08		&92.38	&0.07	&300  \\ \hline
    \hline
    \multicolumn{8}{c}{OGBN-Products} \\ \hline     	&\multicolumn{2}{c|}{\cdzero{}}	&		\multicolumn{2}{c|}{\cdf{}}	&		\multicolumn{2}{c|}{\zeroc{}} & \\ \hline		
\#soc- kets	&Acc.(\%)	&$lr$	&	Acc.(\%)&	$lr$&	Acc.(\%)&	$lr$&	\#epochs \\ \hline
    1	&77.63	&0.01		&77.63	&0.01		&77.63	&0.01	&200  \\
    2	&77.12	&0.05		&77.65	&0.05		&78.42	&0.08	&200  \\ 
    4	&77.35	&0.05		&79.14	&0.07		&78.91	&0.08	&200  \\
    8	&77.49	&0.08		&79.18	&0.07		&79.10	&0.08	&200  \\
    16	&77.47	&0.08		&78.00  &0.08		&78.95	&0.08	&200  \\
    32	&77.45	&0.07		&78.00  &0.08		&78.37	&0.08	&200  \\
    64	&77.25	&0.07		&77.64	&0.08		&77.76	&0.08	&200  \\ \hline
    \hline
    \multicolumn{8}{c}{OGBN-Papers} \\ \hline
    &\multicolumn{2}{c|}{\cdzero{}}	&		\multicolumn{2}{c|}{\cdf{}}	&		\multicolumn{2}{c|}{\zeroc{}} &  \\ \hline		
\#soc- kets	&Acc.(\%)	&$lr$	&	Acc.(\%)&	$lr$&	Acc.(\%)&	$lr$ & \#epohs\\ \hline
    1 & 41.29 & 0.03& 41.29 & 0.03& 41.29 & 0.03 & 200\\
    128 & 37.9 & 0.01 & 37.65 & 0.01 & 36.74 & 0.01 & 200\\ \hline
    \end{tabular}
    \label{tab:accuracy}
\end{table}

\subsubsection*{\textbf{Memory and Communication Analysis}}
In this section, we chart out the memory requirements of the GraphSAGE model and report the actual memory consumption for the OGBN-Papers dataset.
As discussed in Section~\ref{sec-results-exp-setup}, the GraphSAGE model contains three layers; within each layer, the neighborhood aggregation step is followed by the multi-layer perceptron (\MLP{}) operations.
Let $N$, $f$, $h_1$ and $h_2$, and $l$ be the number of partition vertices, features, first hidden layer neurons, second hidden layer neurons, and labels, respectively.
Also, let $w_1$, $w_2$, and $w_3$ be the weight matrices at each of the three layers of the \MLP{}.
Memory required by the GraphSAGE model is as follows.
(1) The weight matrix dimensions, $w_1$: $f\times h_1$, $w_2$: $h_1 \times h_2$, and $w_3$: $h_2\times l$.
(2) The input feature matrix dimensions, $N\times f$.
(3) The neighborhood aggregation output dimensions at each of the three layers,  $N\times f$, $N\times h_1$, and $N\times h_2$.
(4) Similarly, the \MLP{} operation output dimensions at each of the three layers, $N\times h1$, $N\times h2$, and $N\times l$.
The intermediate results at each layer need to be stored to facilitate the backpropagation of the gradients.
Additionally, in a distributed setting, memory is required for buffering the data for communication.
In \cdzero{} and \cdr{} algorithms, the amount of communication per partition is directly proportional to the number of split-vertices in the partitions.
Table~\ref{tab:mem-comm} shows the memory consumption of the distributed algorithms and the percentage of split-vertices per partition for a different number of partitions for OGBN-Papers dataset.

\begin{table}[htb]
    \centering
        \caption{Per epoch peak memory requirements of the distributed algorithms and split-vertices percentage in a partition for OGBN-Papers dataset.}
    \vspace{-10pt}    
    \begin{tabular}{l|r|r|r}
    \hline
         Partitions & $32$ & $64$ & $128$  \\ \hline
         \cdzero{} Memory (GB)& $199$ & $124$ & $78$ \\
         \cdf{} Memory (GB)& $311$    & $196$ & $120$ \\
         \zeroc{} Memory (GB)& $180$  & $112$ & $70$ \\ \hline
         Split-vertices/partition (\%) & $90$ & $92$ & $93$\\ \hline
    \end{tabular}
    \label{tab:mem-comm}
\end{table}

\subsubsection*{\textbf{Comparison with Current Solutions}}

We compare our solution with a recent distributed solution for mini-batch training with neighborhood sampling -- \DistDGL{}. On account of a different aggregation strategy used in \DistDGL{}, we compare our full-batch training approach with it based on total aggregation work and time per epoch. We use the same \GNN{} model architecture as \DistDGL{} for OGBN datasets.
In \DistDGL{}, during each mini-batch computation, the amount of aggregation work varies at each hop, depending on the number of vertices, their average degree (fan-out), and feature vector size.
For full-batch training, the amount of work at each hop varies with feature vector size only.

Out of 2,449,029 vertices of OGBN-Products dataset, 196,615 are labeled training vertices.  Table~\ref{tab:dist-dgl} \& ~\ref{tab:distgnn} show the number of vertices, average degree, and feature vector size per hop.
In \DistDGL{}, with equal distribution of training vertices per socket, each socket processes a roughly equal number of batches.
Table~\ref{tab:dist-dgl} \& ~\ref{tab:distgnn} also highlights the total work that \DistDGL{} and \DistGNN{} do on single and $16$ sockets, in terms of Billions of Ops (B Ops). The total work per hop is calculated as the product of number of vertices, feature size, and average vertex degree. 
Our solution performs $\approx{} 4\times-13\times$ more work ($\approx{} 77.18e^9/19.98e^9$ ops) and ($\approx{} 18.8e^9/1.41e^9$ ops), respectively, per epoch than \DistDGL{} due to complete neighborhood aggregation. However, even with this increase in amount of work, our solution reports comparable or even better epoch time on similar hardware (\DistDGL{} uses either of Intel Xeon Skylake or Cascade Lake \CPU{} with $96$ {\tt VPUs} on AWS instance { \tt m5.24xlarge}) (Table~\ref{tab:comparison}). We see a similar trend with OGBN-Papers dataset which has 1,207,179 training vertices out of 111,059,956 vertices in the graph.


\begin{table}[h]
    \centering
        \caption{Aggregation work done (billion operations) per hop, per mini-batch, and per socket using neighborhood sampling by \DistDGL{}. The mini-batch size is $2000$. Dataset used: OGBN-Products.}
\vspace{-10pt}
    \begin{tabular}{l|r|r|c|r}
    \hline
    \multirow{3}{*}{Hops} & \multirow{3}{*}{\#vertices} & Avg. & \multirow{3}{*}{\#feats} & Total work   \\
         &            & deg. &         &per socket   \\ 
         &        &         &          & (B ops) \\ \hline
    Hop-2 &  233,692 &  5  & 100 & 0.116 \\
    Hop-1 &  30,214  &  10 & 256 &  0.077  \\
    Hop-0 &  2,000   &  15 & 256 &  0.007  \\ \hline
    \multicolumn{4}{l|}{1 Mini-batch }  &  0.202  \\
    \multicolumn{4}{l|}{1 Socket (99 Mini-batches per socket)} & 19.98 \\ 
    \multicolumn{4}{l|}{16 Sockets (7 Mini-batches per socket)} & 1.41 \\  \hline
    \end{tabular}
    \label{tab:dist-dgl}
\end{table}

\begin{table}[h]
    \centering
    \caption{Aggregation work done per hop and per full batch using complete neighborhood aggregation by \DistGNN{}. Here full batch represents a partition and each socket executes one partition. Dataset used: OGBN-Products.}    
    \vspace{-10pt}
    \begin{tabular}{l|l|r|r|c|r}
    \hline
    \#soc-  & \multirow{3}{*}{Hops} & \#vertices/ & Avg.& \multirow{3}{*}{\#feats} & Total work  \\ 
    kets    &  & partition           & deg. &         &per socket \\ 
    & & & & &(B Ops)\\ \hline
    \multirow{4}{*}{1}&Hop-2 &  2,449,029 &  51.5  & 100   &    12.61   \\
    &Hop-1 &  2,449,029      &  51.5 & 256    &   32.29   \\
    &Hop-0 &  2,449,029      &  51.5 & 256 &   32.29   \\ \cline{2-6}
    &Full Batch &  \multicolumn{2}{c}{}&               &   77.19   \\ \hline
    \multirow{4}{*}{16}&Hop-2 &  596,499  &  51.5  & 100   &    3.07    \\
    &Hop-1 &  596,499  &  51.5  & 256   &    7.86    \\
    &Hop-0 &  596,499  &  51.5  & 256   &   7.86    \\ \cline{2-6}
    &Full Batch &         \multicolumn{2}{c}{}&     &   18.80     \\ \hline
    \end{tabular}
    \label{tab:distgnn}
\end{table}

\begin{table}[htb]
     \caption{Training time for \DistDGL{} and \DistGNN{} on OGBN-Products.}
     \vspace{-10pt}
     \centering
     \begin{tabular}{c|c|c}
     \hline
          \#sockets& \DistDGL{} time (s) & \DistGNN{} (\cdf{}) time (s) \\ \hline
          1  & 20 & 11 \\
          16 & 1.5 & 1.9 \\ \hline
     \end{tabular}
     \label{tab:comparison}
 \end{table}
 
CAGNET, a distributed solution to \GNN{} training, performs complete neighborhood aggregation on a \GPU{} cluster. Due to very different cluster configuration and intra- and inter-node network topologies, we do not compare \DistGNN{} with CAGNET. 

%% file: conclusion.tex
\section{Conclusion and Future Work}
\label{sec-conclusion}


In this paper, we present \DistGNN{}, the first ever and a highly efficient distributed solution for full-batch \GNN{} training on Intel Xeon\ \CPU{}s.
The aggregate operation is data-intensive. On a single socket Intel  Xeon \CPU{}, we accelerated it by identifying and optimizing computational bottlenecks. Our distributed solutions employ avoidance and reduction algorithms to mitigate communication bottlenecks in the aggregation operation. To reduce communication volume, we leveraged vertex-cut based graph partitioning and overlapped communication with computation across epochs as delayed partial aggregates. We demonstrated the performance of \DistGNN{} on a set of common benchmark datasets with the largest one having hundred million vertices and over a billion edges. With our delayed partial aggregate algorithms the accuracy is sometimes better than single-socket run and remains within $1\%$ with the increase in number of partitions. \zeroc{} algorithms showing superior accuracy on OGBN-Products datasets points to the clustering effect on the accuracy.

In future work, we expect to demonstrate highly scalable \DistGNN{} for mini-batch training across various datasets. We will further analyze the accuracy of \zeroc{} and \cdr{} algorithms on various datasets and different \GNN{} model architectures that have state-of-the-art accuracy. We also expect to extend \DistGNN{} to different \GNN{} models, beyond GraphSAGE. To further reduce communication volume, we will deploy low-precision data formats such {\tt FP16} and {\tt BFLOAT16}, with convergence using \DistGNN{}.